\documentclass[10pt,twocolumn,letterpaper]{article}

\usepackage{cvpr}
\usepackage{booktabs}
\usepackage{multirow}
\usepackage{array}

\usepackage{algorithm}
\usepackage{algorithmic}
\usepackage{graphicx}
\usepackage{booktabs}
\usepackage[table]{xcolor}
\usepackage{amsmath}
\usepackage[accsupp]{axessibility}  

\newcommand{\joonseok}[1]{{\color{black}#1}}

\definecolor{cvprblue}{rgb}{0.21,0.49,0.74}

\newcommand{\method}{DiVT}
\newcommand{\fullname}{Disentangled Visual Tokenization}

\usepackage[pagebackref,breaklinks,colorlinks,allcolors=cvprblue]{hyperref}

\def\paperID{40601} 
\def\confName{CVPR}
\def\confYear{2026}

\title{A More Word-like Image Tokenization for MLLMs}

\author{Hyun Lee$^1$, Hyemin Jeong$^1$, Yejin Kim$^1$, Hyungwook Choi$^1$\\Hyunsoo Cho$^{2*\dagger}$, Soo Kyung Kim$^{2*\dagger}$, Joonseok Lee$^{1*}$\\
$^1$Seoul National University, $^2$Ewha Womans University\\
{\tt\small \{hyun86,hyeminjeong,a2000yejin,chooi221,joonseok\}@snu.ac.kr, \{chohyunsoo,sookim\}@ewha.ac.kr}
}

\begin{document}
\maketitle

\renewcommand{\thefootnote}{\fnsymbol{footnote}}
\footnotetext[1]{Corresponding authors}
\footnotetext[2]{Department of AI, Institute for Multiscale Matter and Systems}


\begin{abstract}
Modern multimodal large language models (MLLMs) typically keep the language model fixed and train a visual projector that maps the pixels into a sequence of tokens in its embedding space, so that images can be presented in essentially the same form as text. However, the language model has been optimized to operate on discrete, semantically meaningful tokens, while prevailing visual projectors transform an image into a long stream of continuous and highly correlated embeddings. This causes the visual tokens to behave differently from the word-like units that LLMs are originally trained to understand. We propose a novel \fullname\ (\method) that clusters patch embeddings into coherent semantic units, so each token corresponds to a distinct visual concept instead of a rigid grid cell. \method\ further adapts its token budget to image complexity, providing an explicit accuracy-compute trade-off modifying neither the vision encoder nor the language model.
Across diverse multimodal benchmarks, \method\ matches or surpasses
baselines with significantly fewer visual tokens, demonstrating robustness under limited token budgets, significantly reducing memory cost and latency while making visual inputs more compatible with LLMs.
Our code is available at \url{https://github.com/snuviplab/DiVT}.
\end{abstract}

\section{Introduction}
\label{sec:intro}

Large Language Models (LLMs) have shown remarkable capabilities in understanding and generating language through fine-grained textual representations.
Extending it to the visual domain, Multimodal Large Language Models (MLLMs) aim to seamlessly integrate visual information in a similar form with the text tokens, enabling unified multimodal understanding.
To avoid the high computational cost of training from scratch, it is common to adopt a pre-trained LLM for its reasoning ability and a pre-trained vision encoder (\eg CLIP~\cite{clip}, SigLIP~\cite{SigLIP}) to map the pixel-level signals to a semantic latent space.
Since these two pre-trained models operate on different latent spaces, a visual projector is trained to map from one to the other, usually from the visual space to the textual.
Consequently, 
the projector should be able to organize visual information into a token sequence that mirrors the semantic and structural properties of text tokens to fully leverage the LLM's linguistic and reasoning capabilities.


\begin{figure}[t]
    \centering
    
    \begin{subfigure}[b]{0.24\linewidth}
        \centering
        \includegraphics[width=\linewidth]{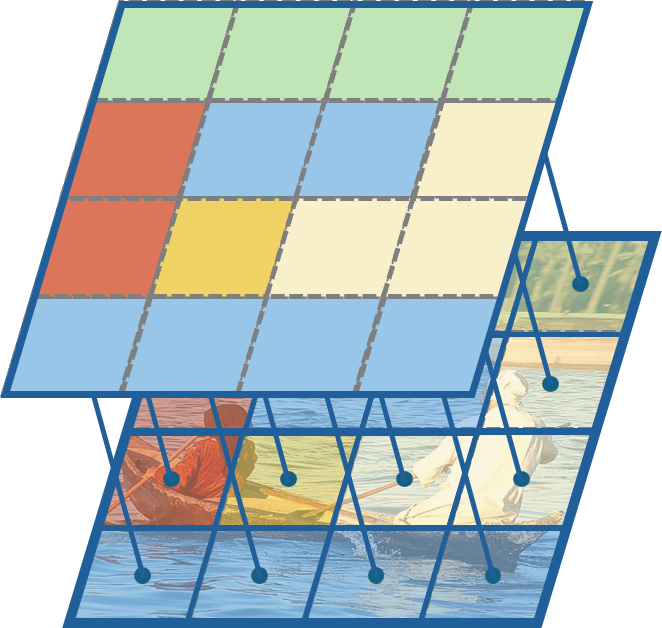}
        \caption{MLP}
        \label{fig:sub_mlp}
    \end{subfigure}
    \hfill
    \begin{subfigure}[b]{0.24\linewidth}
        \centering
        \includegraphics[width=\linewidth]{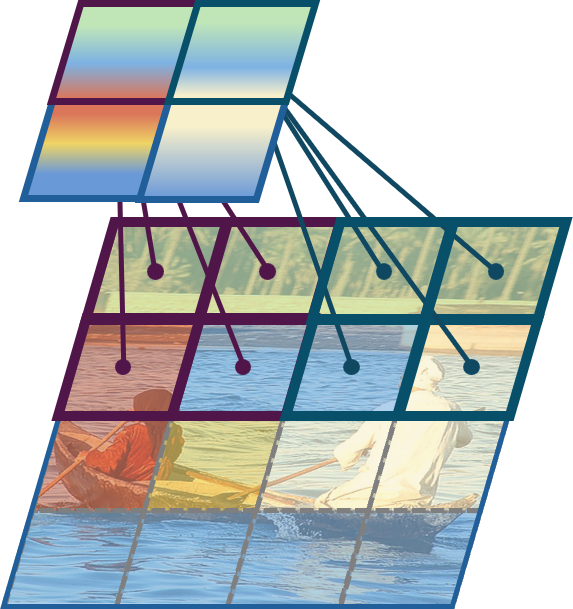}
        \caption{Grid-wise}
        \label{fig:sub_grid}
    \end{subfigure}
    \hfill
    \begin{subfigure}[b]{0.24\linewidth}
        \centering
        \includegraphics[width=\linewidth]{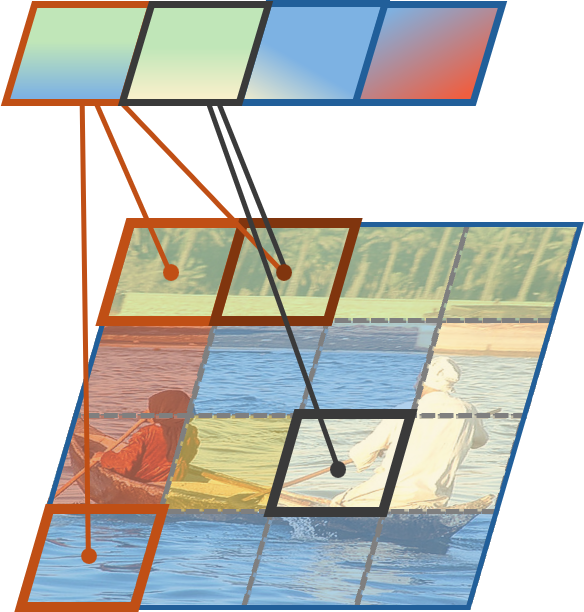}
        \caption{Resampler}
        \label{fig:sub_resampler}
    \end{subfigure}
    \hfill
    \begin{subfigure}[b]{0.24\linewidth}
        \centering
        \includegraphics[width=\linewidth]{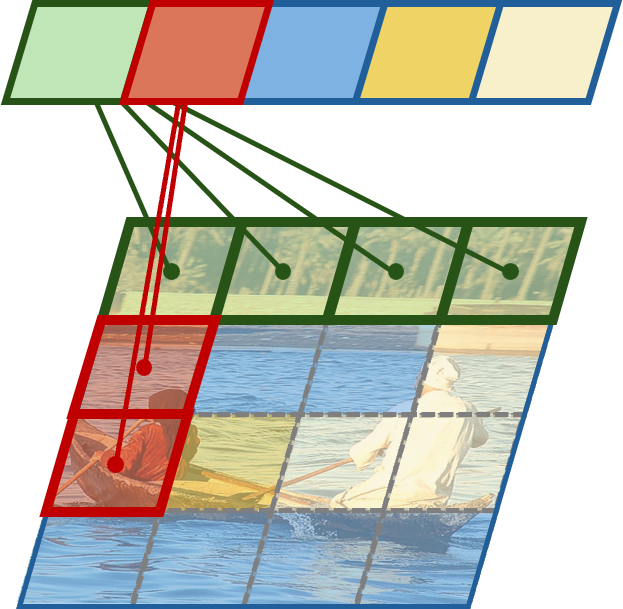}
        \caption{\method\ (Ours)}
        \label{fig:sub_ours}
    \end{subfigure}
    \caption{
    \textbf{Comparison with existing projectors.}
    Patch features (bottom layer) are mapped to visual tokens (top layer). Each color represents the principal semantic of the patch.
    }
    \label{fig:method_overview}
\end{figure}

However, most existing visual projectors are implemented as a simple linear layer~\cite{liu2023visual, shikra} applied to fixed patch features from the vision encoder (\cref{fig:sub_mlp}).
This design adheres to the rigid patchification scheme of ViT~\cite{vit}, which evenly splits an image into a set of fixed-size image patches.
While this approach provides a straightforward way to utilize all the visual information from the encoder, it inevitably introduces significant redundancy and incurs unnecessary computational cost.
To mitigate this redundancy, recent work~\cite{li2025tokenpacker, chu2024mobilevlm, cha2024honeybee, pixel-shuffle, flashsloth} aggregates adjacent patches to form a smaller set of visual tokens (\cref{fig:sub_grid}).
Resampler-based approaches~\cite{li2023blip, dai2023instructblip, qwenvl} generate a compact set of learnable queries that summarize information by globally attending to all visual features, enjoying further flexibility (\cref{fig:sub_resampler}).


We see, however, the following three primary discrepancies between the widely-used visual tokens and text tokens in their formation.
First, \textbf{the patch features are semantically entangled}, not just because they are constructed with a fixed grid without considering their content but also because they have already undergone multiple layers of self-attention in the vision encoder.
Thus, using these raw patch features without explicit semantic disentanglement inevitably propagates such mixed representations into the projector, leading to entangled tokens. 
Text tokens, in contrast, are generated by discrete tokenizers such as Byte Pair Encoding (BPE)~\cite{bytepairencoding}, which segment text into fixed, independent units with limited inter-token interactions.
Thus, their contextual connection usually emerges only after repeated self-attentions in LLMs.

Second, although the amount of information greatly varies across images, existing methods still produce \textbf{a fixed number of visual tokens} for every input.
This often induces an unnecessarily redundant set of tokens for a large portion of patches depicting a single concept (\cref{fig:sub_mlp}), or conversely, a loss of details when the image is forced into fewer tokens than it actually needs.
This contrasts with how a linguistic expression used to describe a scene, where the length of an expression usually varies according to the complexity of the scene (\cref{fig:sub_ours}).

Third, existing approaches \textbf{lack control over the amount of information} encoded within each token.
Whereas textual tokenization offers variable segmentation (\eg sub-words vs.\ whole words) to balance expressiveness and sequence length, current visual tokenization relies on spatial operations with no principled way to control how finely or coarsely an image is partitioned in a manner compatible with LLM-based reasoning.

To address these limitations, we propose a clustering-based visual tokenization approach that aims to explicitly align with LLMs.
Instead of mapping each patch independently or aggregating patches purely by \textit{spatial} proximity, our method clusters patch features from the vision encoder into \textit{semantically} coherent units, with each cluster forming a distinct visual token.
This encourages disentanglement across tokens, so that each token corresponds more closely to a specific visual concept (\eg an object, part, or salient region) rather than a mixed patch of unrelated content.
Crucially, the tokenizer is trained using only the language modeling objective, without any external supervision, so that the resulting visual tokens are shaped by how the LLM internally organizes and exploits semantic information.
Henceforth, we use the terms \textit{projector} and \textit{tokenizer} interchangeably throughout this paper.

Our design incorporates dynamic token allocation; that is, the number of tokens to represent an image is adaptively determined by its content.
The token budget naturally scales with the image's semantic complexity, avoiding both excessive redundancy in simple scenes and insufficient capacity in cluttered ones.
Also, the amount of information per token is controlled by a clustering threshold at training, providing an interpretable knob to adjust how finely or coarsely image regions are grouped to form a token.
Interestingly, this threshold can also be adjusted at inference time, allowing practitioners to trade-off representational detail against memory and latency without retraining, and to match computational budgets in deployment.


We evaluate our 
LLM-friendly visual tokenizer across a broad suite of multimodal benchmarks under varying token budgets.
Empirically, it consistently matches or surpasses baselines with significantly fewer visual tokens, demonstrating robustness under limited token budgets, where the performance gain becomes more evident as the token budget gets more constrained.
The ability to adapt token counts and granularity at inference time yields a practical, training-free mechanism to balance cost and fidelity.
Notably, our approach is agnostic to the choice of vision encoder and remains effective when scaled to larger LLMs, underscoring its practicality and broad applicability for cost-efficient, human-aligned multimodal understanding.


\begin{figure}[t]
    \centering
    \includegraphics[width=0.85\linewidth]{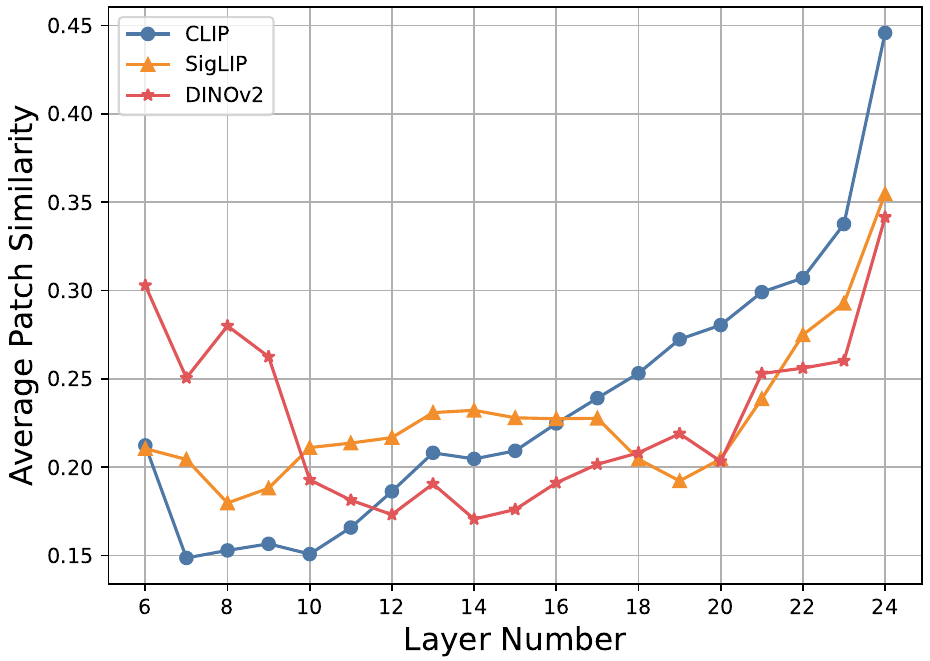}
    \caption{
    \textbf{Patch similarity across ViT layers}.
    Patch-wise cosine similarity increases in deeper layers, indicating that repeated self-attention homogenizes patch embeddings within an image.}
    \label{figure:vit_patch_sim}
    \vspace{-0.2cm}
\end{figure}

\section{Motivation for Redesigning Visual Tokens}
\label{sec:motivation}

We first question if the current visual representations fed into MLLMs behave like language tokens, or collapse to overly similar embeddings within each input.

\begin{figure*}[t]
    \centering
    \includegraphics[width=0.9 \linewidth]{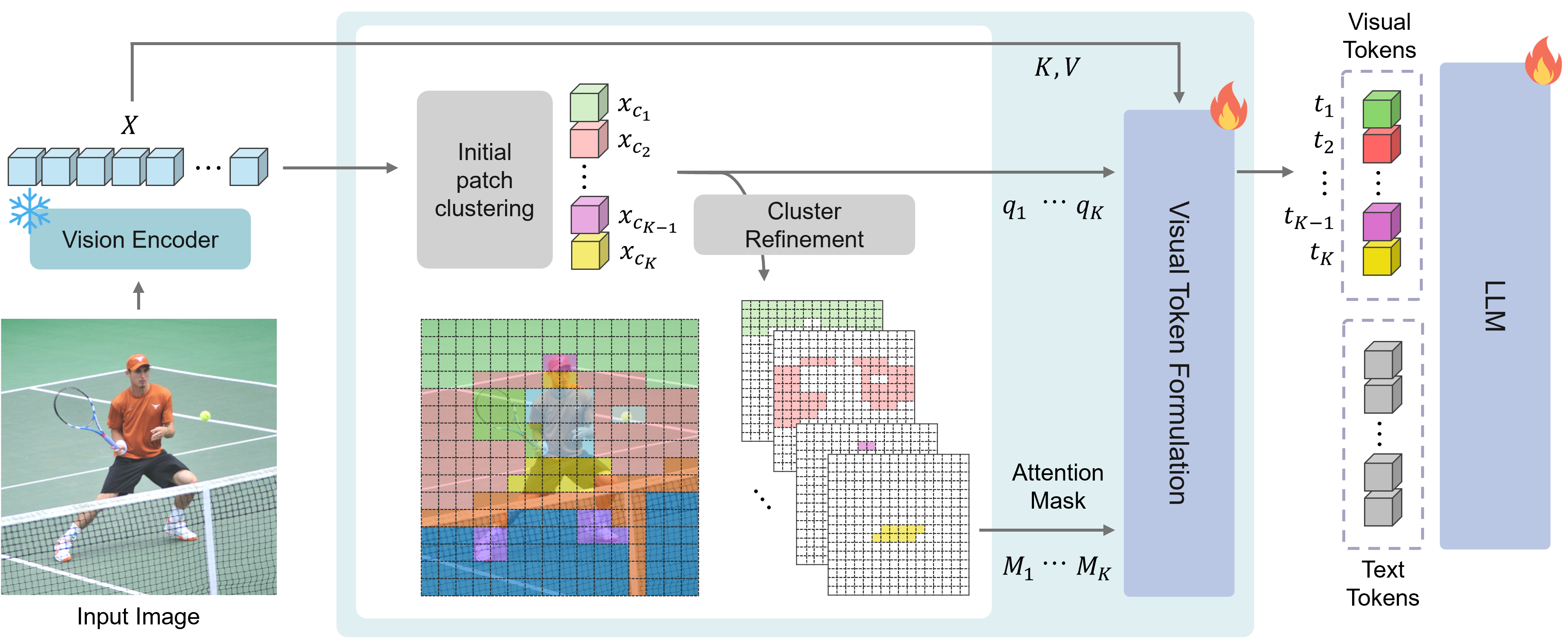}

    \caption{
    \textbf{Overview of \method}. 
    The process consists of three main stages: 
    (1) \textit{Initial patch clustering}, which elects representative patch centroids based on feature diversity (\cref{sec:method:dynamic_alloc}); 
    (2) \textit{Cluster refinement} for semantically more coherent groups (\cref{sec:method:refine}); 
    (3) \textit{Visual token formulation} to aggregate information within each cluster to semantically disentangled visual tokens (\cref{sec:method:token_formulation}).
    }
    \label{fig:visual_token_arch}
    \vspace{-0.5cm}
\end{figure*}

\subsection{Entanglement within Visual Embeddings}
\label{sec:motivation:entangle}

We hypothesize that repeated self-attention and image-level objectives make visual embeddings within an image highly similar, and that a linear projector in \cref{fig:sub_mlp} largely preserves this redundancy in the language model space.
To test this hypothesis, we conduct a toy-scale experiment.
Specifically, we take a CLIP-style vision transformer (ViT) encoder following the standard patchification scheme.
For each image, we produce patch embeddings at different transformer layers using this encoder, and measure cosine similarity between all pairs of patches from the same image, then take the average over pairs and images.
We report the pairwise cosine similarity averaged over all token pairs of 500 images from 
{MMBench~\cite{mmbench}.
\iffalse, separately for language tokens from the LLM and for visual tokens produced by a linear MLP projector.\fi}

\cref{figure:vit_patch_sim} shows that intermediate layers maintain moderate variation among visual embeddings, whereas higher layers exhibit much skyrocketed similarity within each image.
This suggests that visual embeddings become strongly entangled as global context is repeatedly mixed through self-attention, especially under image-level pre-training objectives such as contrastive learning or classification.
As a result, the final visual tokens form a long sequence with many near-duplicate entries, which redundantly inflates the KV cache and spreads attention over redundant evidence.

\subsection{Similarity between visual and language tokens}
\label{sec:motivation:sim}

To quantitatively compare the visual and language sides of the MLLMs, we measure the token similarity within each modality.
Language tokens show significantly lower mutual similarity of $0.0378\pm 0.0002$ on average, consistent with their discrete and well-separated nature.
Visual tokens from the MLP projector reveal significantly higher similarity of $0.3823\pm 0.0018$, verifying the redundancy suggested above.

Overall, these results reveal a structural mismatch inside the current visual-language models.
Vision encoders tend to produce entangled embeddings, and linear projectors preserve this redundancy when mapping them into the language model space.
This motivates a visual tokenization scheme that explicitly targets semantic disentanglement and compact concept-level tokens in the next section.


\section{\fullname}
\label{sec:method}

Motivated by the analysis on the entanglement problem inherent in current visual tokens (\cref{sec:motivation}), we introduce our \fullname\ (\method) designed to produce semantically coherent and disentangled visual tokens, supporting adaptive token lengths for each image.
Unlike the linear projectors that ignore patch entanglement, our approach restructures visual information into \textit{semantic} units, better compatible with the discrete text tokens in LLMs.
We subsequently detail our proposed approach in \cref{fig:visual_token_arch}:
initial patch clustering (\cref{sec:method:dynamic_alloc}), cluster refinement (\cref{sec:method:refine}), visual token formulation (\cref{sec:method:token_formulation}), and semantic granularity control (\cref{sec:method:control}).

\subsection{Initial Patch Clustering}
\label{sec:method:dynamic_alloc}
\label{subsec:centroid_selection}

\begin{figure}[t]
    \vspace{-0.1cm}
    \centering
    \includegraphics[width=\linewidth]{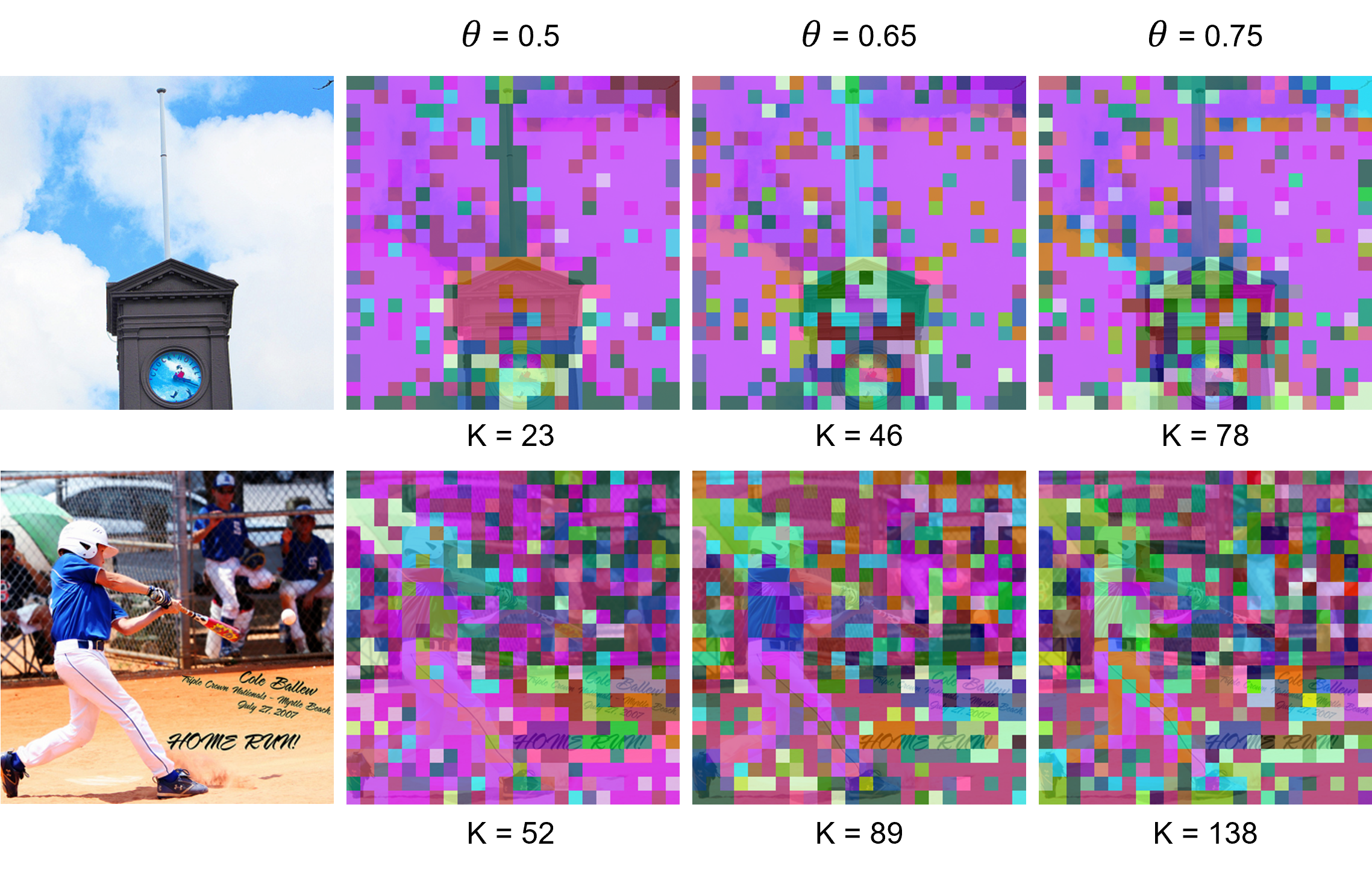}
    \vspace{-0.6cm}
    \caption{\textbf{Illustration of dynamic token clustering}. An image with relatively simpler content (top) uses less number of clusters than one with a more complex scene (bottom). See Appendix \ref{appendix:cluster_vis} for more examples.}
    \label{fig:clustering_visualization}
    \vspace{-0.4cm}
\end{figure}

In order to produce a set of visual tokens that are semantically disentangled and free from redundancy, we begin with clustering the patches based on their features.
A natural option is to use a standard clustering algorithm such as $k$-means, but it requires prefixing the number of clusters $k$, thereby enforcing the same token budget for all images regardless of their content.
This highlights a broader limitation of existing tokenization strategies, including the linear projectors and spatial reduction methods such as strided pooling, pixel-shuffle, or grid-level subsampling.  
All of these methods produce a fixed number of visual tokens \emph{a priori}, rather than letting the token count reflect the semantic complexity of the input image.
Unlike text, where detailed or information-rich sentences naturally expand into a longer sequence, fixed-budget visual tokenization cannot allocate more tokens to scenes containing richer content.

To overcome this structural rigidity, we aim to dynamically decide the number of clusters based on
the feature distribution itself.
Our design is guided by the observation that patches whose feature vectors lie in dense regions of the embedding space, \textit{i.e.}, have many neighbors above a cosine-similarity threshold, tend to correspond to semantically dominant structures, whereas patches in sparse regions typically encode fine-grained or isolated details.
Instead of imposing a predetermined token budget, we thereby derive cluster centroids directly from the similarity structure of the patch embeddings.

\joonseok{Formally, we denote the set of patch features by $\mathcal{X}=\{\mathbf{x}_i\}_{i=1}^{N}$, where $N$ is the number of patches per image.
With a ViT-like vision encoder, $N$ is usually fixed regardless of the image content.}
Given $\mathcal{X}$, we compute the patch-wise cosine similarity matrix $\mathbf{S} \in \mathbb{R}^{N \times N}$.
\joonseok{We define the patches whose pairwise similarity exceeds some threshold $\theta$ as \emph{neighbors}.
That is, taking $\mathbf{S}$ only with its elements above $\theta$ as the adjacency matrix would create a neighborhood graph among the patches.
In this graph, we greedily select a node (patch) with the largest degree (the number of edges connected to it).
The chosen node $\mathbf{x}$ becomes the centroid of the first cluster $c_1$, denoted by $\mathbf{x}_{c_1}$, and all neighboring nodes to it construct the cluster $c_1$.
We remove all nodes belonging to $c_1$, and repeat this process to select the subsequent clusters $c_k$ for $k = 2, 3, ..., K$, until no node remains.
In this way,} the remaining candidates are at least $\theta$ apart from any previously chosen centroids.
This guarantees that \joonseok{subsequently chosen centroids are semantically distinct from previously selected ones}, rather than redundant variations of the same structure.
\cref{alg:adaptive_centroid} summarizes these steps.
While the algorithm induces an implicit ordering of centroids, we simply arrange them based on their spatial coordinates to form a sequence for LLM input.

This clustering approach satisfies two desirable properties.
First, \joonseok{the number of clusters is adaptively decided based on the image content.}
Images with richer visual content naturally yield \joonseok{a larger number of clusters}, while simpler or homogeneous images yield fewer, as illustrated in \cref{fig:clustering_visualization}.
The resulting token budget therefore adapts to the intrinsic complexity of the scene.
\joonseok{Second, fine-grained details are not discarded.}
Patches that rarely resemble others form isolated \joonseok{clusters}, ensuring that outliers, subtle edges, and small distinctive objects are preserved rather than absorbed into \joonseok{another} broader cluster.
Through this mechanism, the visual token count is determined adaptively by the feature distribution, providing a natural basis for controllable token granularity discussed in \cref{sec:method:control}.

\begin{algorithm}[t]
\small
\caption{Adaptive Centroid Selection}
\label{alg:adaptive_centroid}
\textbf{Input:} Patch features $\mathcal{X}=\{\mathbf{x}_i\}_{i=1}^N$, similarity threshold $\theta$, $\mathbb{I}(\cdot)$ is an indicator function.\\
\textbf{Output:} Centroid indices $\mathcal{C}$

\begin{algorithmic}
\STATE $\mathbf{S}_{i,j} = \text{cos\_sim}(\mathbf{x}_i, \mathbf{x}_j)$
\STATE $n_i = \sum_j \mathbb{I}(\mathbf{S}_{i,j}>\theta)$ 
\STATE candidates = argsort$(n, \text{desc})$
\STATE $\mathcal{C} = [\;]$
\WHILE{candidates $\neq \emptyset$}
    \STATE $c = \text{candidates}[0]$
    \STATE $\mathcal{C}.append(c)$
    \STATE candidates = candidates $\setminus \{j \mid \mathbb{I}(\mathbf{S}_{c,j}>\theta)\}$
\ENDWHILE
\RETURN $\mathcal{C}$
\end{algorithmic}
\end{algorithm}

\subsection{Cluster Refinement for Disentanglement}
\label{sec:method:refine}

\joonseok{Although we have generated the initial clusters in \cref{subsec:centroid_selection}, the cluster allocation can be far from optimum due to its greedy nature.
To be specific, an image patch may belong to more than one cluster (that is, it is close enough to more than one centroid patch).
According to \cref{alg:adaptive_centroid}, its initial cluster would be the one having more neighbors, but it would be ideal to associate it with the closest centroid, which has been discovered in the later steps.
For this reason, we refine the cluster assignment for better semantic disentanglement across the clusters.}

Given patch features $\mathcal{X}=\{\mathbf{x}_i\}_{i=1}^{N}$ and \joonseok{their corresponding} centroid indices $\{c_k\}_{k=1}^{K}$ constructed in \cref{subsec:centroid_selection}, each patch is assigned to the centroid with the highest similarity:
\begin{equation}
    \mathcal{C}_k = \{\, \mathbf{x}_i \mid k =  \underset{j}{\text{argmax}} \cos(\mathbf{x}_i, \mathbf{x}_{c_j}) \,\},
\end{equation}
\joonseok{where $\mathbf{x}_{c_j}$ indicates the centroid of the cluster $j$.}
\joonseok{Each resulting $\mathcal{C}_k$ for $k = 1, ..., K$ represents} a set of patches that share coherent semantics in the feature space.
\joonseok{In this way, the initial clustering step in \cref{sec:method:dynamic_alloc} serves only for selecting the set of centroids (roughly speaking, this can be seen as a discrete density estimation), while this refinement step achieves the desired disentanglement.}

\subsection{Visual Token Formulation}
\label{sec:method:token_formulation}
Once the clusters are finalized in \cref{sec:method:refine}, we \joonseok{aggregate information from the patches in each cluster to produce a single visual token per cluster.
Specifically, we adopt cross-attention using the centroid as the query:}
\begin{equation}
    \mathbf{Q}_k = \mathbf{W}^Q \mathbf{x}_{c_k}, \
    \mathbf{K}_i = \mathbf{W}^K \mathbf{x}_i, \
    \mathbf{V}_i = \mathbf{W}^V(\mathbf{x}_i + \mathbf{P}_i),
\end{equation}
where \joonseok{$\mathbf{W}^{\{Q,K,V\}}$ are learnable parameters, and}
$\mathbf{P}_i$ is a learnable positional embedding that provides spatial context.
We inject $\mathbf{P}_i$ only into the value branch because it directly influences the aggregated token content, whereas adding it to $\mathbf{Q}$ or $\mathbf{K}$ would merely perturb attention scores without providing meaningful structural cues.

To ensure each token to attend inside its cluster only, we apply a cluster-restricted attention mask:
\begin{equation}
    \mathbf{M}_{k,i}=
    \begin{cases}
        0, & \text{if} \ i\in\mathcal{C}_k,\\
        -\infty, & \text{otherwise}.
    \end{cases}
\end{equation}
The resulting visual token $\mathbf{t}_k$ for the cluster $k$ is obtained by
\begin{equation}
    \mathbf{t}_k = \mathrm{MLP}\!\left(
    \sum_{i}\mathrm{softmax}(\mathbf{Q}_k \mathbf{K}_i^\top + \mathbf{M}_{k,i}) \mathbf{V}_i
    \right).
\end{equation}
This design has two key advantages.
First, because the aggregation is restricted to the patches that share relevance with the centroid, the resulting token naturally encapsulates a well-separated semantic unit.
Second, every patch belongs to one cluster, ensuring that no visual content is discarded.
Even fine-grained details (\eg, small objects, object parts, or text characters) contribute to some token, preventing information loss.
Putting them together, these properties enable the DiVT to compress images into a compact token set while preserving patch-level information.

  


\subsection{Controlling Semantic Granularity}
\label{sec:method:control}

The similarity threshold $\theta$ serves as a principled means to control the semantic granularity of the resulting visual tokens.
Since $\theta$ defines which patches are considered neighbors, it implicitly determines the scale at which visual structures are grouped.
A higher $\theta$ enforces stricter grouping criteria, \joonseok{producing a larger number of finer-grained tokens}
that capture subtle variations across the image.
\joonseok{It tends to preserve a more} detailed structure and form smaller semantic units, at the cost of higher token counts and computational cost.
Conversely, a lower $\theta$ encourages broader grouping, generating fewer clusters and more coarse-grained tokens. 
Such representations remain compact, but may lose some localized details, merging visually distinct \joonseok{patches even when their similarity is not extremely high.}

This behavior mirrors the continuum of textual tokenization level, from character-level to subword- and word-level, to balance expressiveness, semantic specificity, and sequence length.
Finer granularity supports detail-oriented representations but leads to longer sequences, while coarser granularity improves efficiency at the risk of under-representing subtle information. 
By adjusting $\theta$, practitioners can flexibly navigate this trade-off to match the needs for the downstream task or computational budget. 
Notably, $\theta$ can also be adjusted \emph{training-free} at inference time, allowing the model to reduce the number of tokens and consequently lower the inference cost without retraining
(see \cref{sec:exp:ablation} for details).
\section{Experiments}
\label{sec:exp}

\begin{table}[t]
    \centering
    \setlength{\tabcolsep}{1.5pt}
    \resizebox{0.5\textwidth}{!}{
        \begin{tabular}{lccccccccc}
            \toprule
            \textbf{Method} & \textbf{\# Tokens} & \textbf{MMB} & \textbf{VQA\textsuperscript{v2}} & \textbf{GQA} & \textbf{MME} & \textbf{MM-Vet} & \textbf{VQA\textsuperscript{Text}} & \textbf{SQA\textsuperscript{IMG}} & \textbf{POPE} \\

            \midrule
            MLP & 576 & 64.3 & 78.5 & 62.0 & 1510.7 & 31.1 & 58.2 & 66.8 & 85.6 \\
            \midrule

            ToME* & 128 & 53.3 & 63.0 & 52.4 & 1088.4 & 27.2 & 49.1 & 59.6 & 62.8 \\
            FastV* & 128 & 56.1 & 61.8 & 49.6 & 1208.9 & 28.1 & 50.6 & 60.2 & 59.6 \\
            PruMerge+* & 128 & 61.3 & 74.7 & 57.8 & 1420.5 & 28.7 & 54.3 & 67.6 & 81.5 \\
            VisionZip* & 128 & 62.0 & 75.6 & 57.6 & 1432.4 & 32.6 & 56.8 & 68.9 & 83.2 \\
            VisPruner* & 128 & 62.7 & 75.8 & 58.2 & \underline{1461.4} & \textbf{33.7} & 57.0 & \underline{69.1} & 84.6 \\
            ATP-LLaVA* & 144\textsuperscript{\dag} & \underline{66.0} & 76.4 & 59.5 & \textbf{1473.9} & - & - & \underline{69.1} & 84.2 \\
            TokenPacker & 144 & 65.1 & \underline{77.9} & \underline{61.9} & - & \underline{33.0} & \underline{57.2} & - & \textbf{87.0} \\
            \method$_{\ \theta={0.75}}$ & 136.5\textsuperscript{\dag} & \textbf{66.7} & \textbf{78.2} & \textbf{62.0} & 1457.6 & 30.2 & \textbf{57.7} & \textbf{70.0} & \underline{86.2} \\

            \midrule

            ToME* & 64 & 43.7 & 57.1 & 48.6 & 922.3 & 24.1 & 45.3 & 50.0 & 52.5 \\
            FastV* & 64 & 48.0 & 55.0 & 46.1 & 1019.6 & 25.8 & 47.8 & 51.1 & 48.0 \\
            PruMerge+* & 64 & 59.3 & 67.4 & 54.9 & 1198.2 & 25.9 & 53.0 & 68.6 & 77.4 \\
            VisionZip* & 64 & 60.1 & 72.4 & 55.1 & 1365.6 & 31.7 & 55.5 & 69.0 & 77.0 \\
            VisPruner* & 64 & 61.3 & 72.7 & 55.4 & 1369.9 & \textbf{32.3} & 55.8 & \underline{69.1} & 80.4 \\
            ATP-LLaVA* & 88\textsuperscript{\dag} & \underline{64.7} & 73.3 & 56.8 & 1401.5 & - & - & 67.2 & 82.6 \\
            TokenPacker & 64 & 64.1 & \underline{77.2} & 61.1 & - & 31.7 & - & - & \textbf{86.3} \\
            \method$_{\ \theta={0.65}}$ & 74.1\textsuperscript{\dag} & \textbf{65.5} & \textbf{77.7} & \underline{61.4} & \textbf{1465.7} & \underline{32.1} & \textbf{57.2} & 68.1 & 85.8 \\
            \method$_{\ \theta={0.62}}$ & 63.7\textsuperscript{\dag} & 64.3 & \textbf{77.7} & \textbf{61.6} & \underline{1463.0} & 30.6 & \underline{57.0} & \textbf{70.6} & \underline{86.2} \\

            \midrule

            ToME* & 32 & 31.6 & 46.8 & 43.6 & 828.4 & 17.3 & 38.3 & 41.4 & 39.0 \\
            FastV* & 32 & 37.8 & 43.4 & 41.5 & 884.6 & 20.7 & 42.5 & 42.6 & 32.5 \\
            PruMerge+* & 32 & 56.8 & 54.9 & 51.1 & 940.8 & 21.4 & 50.6 & 68.5 & 70.9 \\
            VisionZip* & 32 & 57.7 & 67.1 & 51.8 & 1247.4 & 25.5 & 53.1 & \underline{68.8} & 68.7 \\
            VisPruner* & 32 & 58.4 & 67.7 & 52.2 & \underline{1271.0} & 28.8 & \underline{53.9} & \textbf{69.2} & 72.7 \\
            TokenPacker & 36 & \underline{62.8} & \underline{75.0} & \underline{59.6} & - & \underline{29.6} & - & - & \textbf{86.2} \\
            \method$_{\ \theta={0.5}}$ & 35.7\textsuperscript{\dag} & \textbf{65.0} & \textbf{77.0} & \textbf{60.6} & \textbf{1458.2} & \textbf{31.7} & \textbf{57.1} & 68.2 & \underline{85.8} \\
            \midrule
            \method$_{\ \theta={0.4}}$ & 22.4\textsuperscript{\dag} & 64.7 & 76.4 & 60.1 & 1450.9 & 31.7 & 56.1 & 69.1 & 84.8 \\
            \method$_{\ \theta={0.3}}$ & 13.5\textsuperscript{\dag} & 64.2 & 75.3 & 59.2 & 1462.8 & 28.0 & 55.4 & 69.4 & 84.3 \\
            \bottomrule
        \end{tabular}
        }
        \vspace{-0.2cm}
        \caption{\textbf{Comparison with token compression methods on LLaVA-1.5 7B}~\cite{liu2023visual}\textbf{.} 
        Training-free methods are marked with *. 
        \dag denotes that the number of tokens is averaged across the test set, as each sample uses varied number of tokens.
        Token counts for ATP-LLaVA are brought from its original paper~\cite{ATP-LLaVA}.
        \textbf{Bold} and \underline{underline} indicate the best and second-best results, respectively.}
    \label{tab:main_results}
    \vspace{-0.3cm}
\end{table}

We conduct extensive experiments to verify the effectiveness of our method on a diverse set of benchmarks designed to test various capabilities.

\subsection{Experimental Setup}
\label{sec:exp:setting}


\textbf{Benchmarks.}
We evaluate on eight widely-used benchmarks: general conversation and VQA (MMBench~\cite{mmbench}, VQAv2~\cite{vqav2}, GQA~\cite{gqa}, MME~\cite{mme}, MM-Vet~\cite{mm-vet}), OCR-related tasks (TextVQA~\cite{textvqa}), knowledge-based tasks (SQA-IMG~\cite{sqa-img}), and specialized tasks measuring object hallucination (POPE~\cite{pope}).
A higher score indicates better performance across all benchmarks.

\vspace{0.1cm} \noindent
\textbf{Baselines.}
We compare our DiVT against two categories of baselines: training-free token reduction methods, including ToME~\cite{ToME}, FastV~\cite{chen2024fastv}, PruMerge~\cite{prumerge}, VisionZip~\cite{yang2025visionzip}, VisPruner~\cite{vispruner}, and ATP-LLaVA~\cite{ATP-LLaVA}, and TokenPacker~\cite{li2025tokenpacker} which requires training a modified projector.

To specifically isolate the contribution of our projector itself, we further compare DiVT against other prominent visual projectors (\textit{e.g.}, Resampler~\cite{qwenvl}, C-Abstractor~\cite{cha2024honeybee}), originally proposed as a component of various MLLMs.

\vspace{0.1cm} \noindent
\textbf{Implementation Details.}
We replace the MLP projector in LLaVA-1.5 with our proposed projector, leaving all other experimental settings including the model backbone and training datasets unchanged.
We cross-validate $\theta \in \{0.5, 0.65, 0.75\}$,
which correspond to approximately 35.7, 74.1, and 136.5 tokens, respectively.
Additionally, we include $\theta=0.62$ (63.7 tokens) for a fairer comparison with baselines operating at a 64-token budget, and $\theta=0.3$ and $0.4$ (13.5 and 22.4 tokens, respectively) to examine performance under a more aggressively limited budget regime.
See Appendix \ref{appendix:implementation} for more details.

\subsection{Main Results}
\label{sec:exp:main_result}


\textbf{Comparison with Baselines.}
\joonseok{\cref{tab:main_results} compares the performance of competing methods using a fixed LLaVA-1.5 7B backbone.}
For methods generating a variable number of tokens specific to each image, ATP-LLaVA and \method, we report the average token count measured at evaluation.

First of all, our method achieves highly competitive performance compared to the full LLaVA (576 tokens) despite the drastically reduced number of tokens (136.5 to 22.4, depending on $\theta$).
For instance, with 136.5 tokens, our method achieves 66.7 on MMBench and 78.2 on VQAv2, achieving comparable or even superior performance, compared to the full model's 64.3 and 78.5, respectively.

While most other methods are training-free (marked with *), TokenPacker is the only other approach that trains its projector. This direct comparison reveals that \method\ demonstrates superior performance, especially on general conversation and VQA tasks (\textit{e.g.}, MMB, VQAv2 or GQA). This suggests that our approach of organizing tokens based on semantic content is more effective than methods relying on static grid-based compression.

Notably, the performance gain with our projector gets more pronounced with a tighter token budget. 
\method$_{\ \theta=0.5}$, using 35.7 tokens, achieves 65.0 on MMB and 77.0 on VQAv2, substantially outperforming TokenPacker's 36-token model, which scored 62.8 and 75.0, respectively. This advantage consistently holds across GQA (60.6 \textit{vs.}~59.6) and MM-Vet (31.7 \textit{vs.}~29.6).
Furthermore, our most compact models (\method$_{\ \theta=0.3}$, \method$_{\ \theta=0.4}$) emphatically prove this; despite using only 13.5 to 22.4 tokens, they still outperform other baselines on majority of benchmarks.
Grid-based methods such as TokenPacker or token reduction approaches focus on \textit{compressing} tokens and tend to lose core information under limited token budgets, while our DiVT controls the \textit{semantic granularity} of each token, resulting in robust performance with more informative visual tokens.

\begin{table}
    \centering
    \setlength{\tabcolsep}{4pt}
    \resizebox{\linewidth}{!}{
    \begin{tabular}{lcccccc}
        \toprule
        \textbf{Projector} & \textbf{\# Tokens} & \textbf{MMB} & \textbf{VQA}\textsuperscript{v2} & \textbf{GQA} & \textbf{MM-Vet} & \textbf{POPE} \\
        \midrule
        MLP (full)                              & 576 & 64.3 & 78.5 & 62.0 & 31.1 & 85.9 \\
        \midrule
        Average-Pooling                         & 64  & 62.4 & 72.6 & 58.8 & 27.1 & 85.4 \\
        Resampler~\cite{qwenvl}              & 64  & 63.4 & 74.1 & 57.7 & 29.2 & 83.4 \\
        C-Abstractor~\cite{cha2024honeybee}     & 64  & 62.5 & 74.4 & 59.3 & 29.0 & 85.0 \\
        Pixel-Shuffle~\cite{pixel-shuffle}      & 64  & 63.2 & 74.6 & 59.1 & 28.5 & 85.2 \\
        LDP-v2~\cite{chu2024mobilevlm}          & 64  & 63.7 & 75.3 & 59.7 & 30.0 & 85.5 \\
        TokenPacker~\cite{li2025tokenpacker}    & 64  & \underline{64.1} & \underline{77.2} & \underline{61.1} & \textbf{31.7} & \textbf{86.3} \\
        \method\ (Ours)                         & 63.7  & \textbf{64.3} & \textbf{77.7} & \textbf{61.6} & \underline{30.6} & \underline{86.2} \\
        \bottomrule
    \end{tabular}
    }
    \vspace{-0.2cm}
    \caption{\textbf{Comparison of different visual projectors.}}
    \label{tab:projector_comparison}
    \vspace{-0.3cm}
\end{table}

\vspace{0.1cm} \noindent
\textbf{Comparison with Visual Projectors.}
\cref{tab:projector_comparison} compares the performance of various visual projectors, following the experimental setup in TokenPacker~\cite{li2025tokenpacker} with LLaVA-1.5 7B.
Our method consistently achieves the highest scores on MMBench, VQAv2, and GQA, while remaining competitive on MM-Vet and POPE.
This demonstrates that \method\ effectively distills the most important visual information into a compact token set, preserving model performance even with a reduced token budget.
Overall, these results highlight the efficiency and effectiveness of our visual projector compared to existing approaches.

\begin{figure*}
    \centering
    \includegraphics[width=0.8\linewidth]{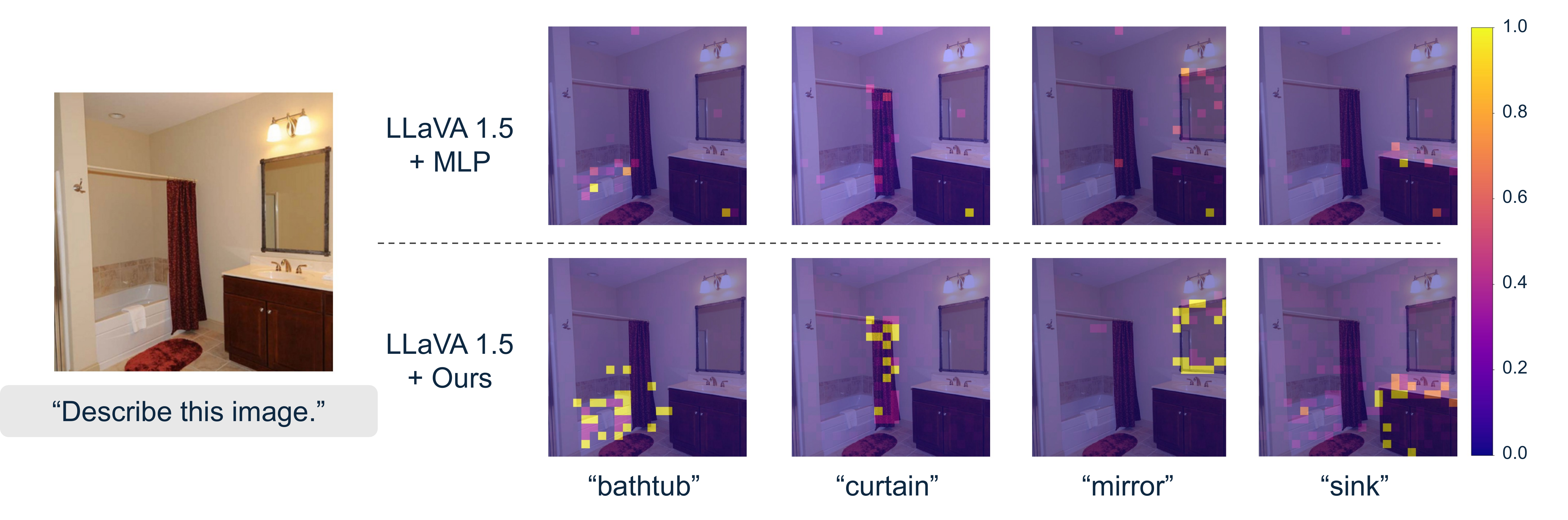}
    \caption{
    \textbf{Qualitative demonstration.} Attention maps highlight the regions in the image that the model attends to for specific object tokens. Our method produces attention clusters tightly focused on the object token, yielding more interpretable patterns, while the MLP projector exhibits more scattered attention over irrelevant regions. See Appendix \ref{appendix:appendix_attention} for more examples.}
    \label{fig:attention_map}
    \vspace{-0.4cm}
\end{figure*}

\vspace{0.1cm} \noindent
\textbf{Qualitative Analysis.}
\cref{fig:attention_map} illustrates the attention map produced by the MLP projector (top) and \method (bottom) for a prompt ``Describe this image''.
Each set of attention maps highlights some regions in the image that the model attends to for a specific word token within its generated caption.

Our method consistently demonstrates more focused and semantically coherent attention, clearly concentrating on the corresponding object.
This sharply contrasts with the MLP projector, which exhibits a more dispersed and less interpretable attention pattern.
These results underscore that our DiVT effectively disentangles visual information into meaningful semantic units, leading to improved interpretability and object grounding.

\subsection{Further Analysis}
\label{sec:exp:ablation}

\textbf{Encoder-Agnostic Versatility.}
A key strength of DiVT is its design as an encoder-agnostic component.
We validate this by applying DiVT to diverse vision encoders, including CLIP~\cite{clip}, SigLIP~\cite{SigLIP}, and DINOv2~\cite{Dinov2}.
We set the threshold to $\theta = 0.6$ for SigLIP and $\theta = 0.66$ for DINOv2, so their average token counts remain comparable to CLIP.

According to the results presented in \cref{tab:backbone_comparison}, our method consistently proves its general applicability, sufficiently preserving the encoders' inherent capabilities with a significantly reduced number of tokens.
Notably, despite using about 13\% of the tokens of the full MLP projector, our method achieves highly comparable performance across all encoders.
This demonstrates that the significant gains in computational efficiency with aggressive token reduction far outweigh the marginal, task-specific performance trade-offs.
This strongly demonstrates that our DiVT does not overfit to a specific feature space but generalizes effectively, confirming its robustness as a plug-and-play module.

\vspace{0.1cm} \noindent
\textbf{Scalability to Larger LLMs.} 
We further examine the scalability of DiVT by integrating it with a larger LLaVA-1.5 13B model (based on Vicuna-13B~\cite{vicuna2023}).
The results in \cref{tab:backbone_comparison} imply that 
the benefit of the semantic disentanglement in visual tokens is robustly maintained and scales effectively with a more capable LLM.

\begin{table}
    \centering
    \setlength{\tabcolsep}{2pt}
    \resizebox{\linewidth}{!}{%
        \begin{tabular}{l * {9}{c}}
            \toprule
            \textbf{Backbone} & \textbf{\# Tokens} & \textbf{MMB} & \textbf{VQA\textsuperscript{v2}} & \textbf{GQA} & \textbf{MME} & \textbf{MM-Vet} & \textbf{VQA\textsuperscript{Text}} & \textbf{SQA\textsuperscript{IMG}} & \textbf{POPE} \\
            \midrule
            CLIP                & 576   & 64.3 & 78.5 & 62.0 & 1510.7 & 31.1 & 58.2 & 66.8 & 85.6 \\
            $\ \ \ $+ \method   & 74.1  & 65.5 & 77.7 & 61.4 & 1465.7 & 32.1 & 57.2 & 68.1 & 85.8 \\
            \midrule
            SigLIP              & 576   & 66.6 & 80.0 & 62.9 & 1498.3 & 33.7 & 60.3 & 69.5 & 85.3 \\
            $\ \ \ $+ \method   & 74.3  & 66.4 & 78.5 & 62.1 & 1506.6 & 31.5 & 58.1 & 68.1 & 84.5 \\
            \midrule
            DINOv2              & 576   & 59.9 & 75.5 & 62.1 & 1266.1 & 25.0 & 46.6 & 66.8 & 85.6 \\
            $\ \ \ $+ \method   & 70.9  & 56.9 & 74.2 & 61.0 & 1362.8 & 23.9 & 46.6 & 65.3 & 85.0 \\
            \midrule
            Vicuna-13B          & 576   & 67.7 & 80.0 & 63.3 & 1531.3 & 36.1 & 61.3 & 71.6 & 85.9 \\
            $\ \ \ $+ \method   & 74.1  & 68.4 & 78.8 & 62.4 & 1491.9 & 34.3 & 59.3 & 70.7 & 86.2 \\
            \bottomrule
        \end{tabular}
        }
    \vspace{-0.2cm}
    \caption{\textbf{Performance comparison with various backbones.}
    \iffalse All results are on LLaVA-1.5 7B.\fi} 
    \label{tab:backbone_comparison}
    \vspace{-0.3cm}
\end{table}



\vspace{0.1cm} \noindent
\textbf{Training-Free Token Adjustment.}
Recall from \cref{sec:method:control} that our DiVT allows training-free control of token numbers by adjusting the similarity threshold.
When a lower $\theta$ is used at inference, the number of resulting tokens would be reduced, increasing the amount of information for each token.

\cref{tab:training_free}, reporting the performance when a different $\theta$ is used from the one used at training (0.65 and 0.75, respectively), indicates that our DiVT trained with a larger $\theta$ can be used with a coarser granularity with minimal performance loss.
Although adjusting the threshold slightly affects the consistency of the semantic granularity of each visual tokens, it provides a flexibility to balance between efficiency and performance.
In contrast, increasing the threshold does not further improve performance.
This is expected, since the tokens have no way to have more fine-grained information than it has been trained on.

We further observe that the model trained with a randomly sampled $\theta$ exhibits consistently stable performance across various inference thresholds, demonstrating robustness to varied token granularities.
However, models trained with a fixed $\theta$ tend to achieve the best performance when evaluated at their corresponding training threshold, suggesting that specialized training for a specific granularity remains optimal.

\begin{table}
    \centering
    \setlength{\tabcolsep}{2pt}
    \resizebox{\linewidth}{!}{%
        \begin{tabular}{l * {9}{c}}
            \toprule
            \textbf{$\theta$} & \textbf{\# Tokens} & \textbf{MMB} & \textbf{VQA\textsuperscript{v2}}  & \textbf{GQA} & \textbf{MME} & \textbf{MM-Vet} & \textbf{VQA\textsuperscript{Text}}& \textbf{SQA\textsuperscript{IMG}} & \textbf{POPE} \\
            \midrule
            0.5             & 35.7 & 63.8 & 76.8 & 60.3 & 1462.7 & 30.6 & 56.7 & 68.2 & 82.7 \\
            \textbf{0.65}   & 74.1 & 65.5 & 77.7 & 61.4 & 1465.7 & 32.1 & 57.2 & 68.1 & 85.8 \\
            0.75            & 136.5& 65.1 & 77.8 & 61.0 & 1465.8 & 30.3 & 57.5 & 69.1 & 83.9 \\
            \midrule
            0.5             & 35.7 & 65.3 & 76.2 & 60.1 & 1429.8 & 31.9 & 56.1 & 69.6 & 83.3 \\
            0.65            & 74.1 & 65.8 & 77.7 & 61.7 & 1479.0 & 32.1 & 57.0 & 69.3 & 85.4 \\
            \textbf{0.75}   & 136.5& 66.7 & 78.2 & 62.0 & 1457.6 & 30.2 & 57.7 & 70.0 & 86.2 \\
            \midrule
            \textbf{0.5}    & 35.7 & 65.5 & 76.9 & 60.6 & 1456.1 & 32.2 & 56.5 & 69.9 & 85.5 \\
            \textbf{0.65}   & 74.1 & 65.6 & 77.6 & 61.1 & 1482.6 & 32.4 & 57.0 & 70.0 & 86.2 \\
            \textbf{0.75}   & 136.5& 65.9 & 77.8 & 62.4 & 1494.5 & 31.9 & 57.0 & 69.9 & 86.4 \\
            \bottomrule
        \end{tabular}
    }
    \vspace{-0.2cm}
    \caption{\textbf{Performance of DiVT with varied similarity threshold at inference}. The original models are trained with $\theta=0.65$, $\theta=0.75$, and a randomly sampled $\theta \in \{0.5, 0.65, 0.75\}$, respectively.}
    \label{tab:training_free}
    \vspace{-0.3cm}
\end{table}

\vspace{0.1cm} \noindent
\textbf{Adaptively-decided Token Counts.}
Since our tokenizer adaptively determines the number of tokens based on the visual complexity of each image, the resulting token count naturally varies across benchmarks.
\cref{tab:token_per_benchmark} reports the average number of tokens for each benchmark at $\theta=0.65$.
Benchmarks containing visually simple images (\eg SQA-IMG) tend to require fewer tokens, whereas those involving more complex or text-heavy scenes (\eg POPE, TextVQA) result in a significantly higher number of tokens.
This variation reflects the intended behavior of our method, which assigns more tokens only when the visual content genuinely calls for them. See Appendix \ref{appendix:appendix_token_stats} for more details.

\begin{table}
    \centering
    \setlength{\tabcolsep}{2pt}
    \resizebox{\linewidth}{!}{
    \begin{tabular}{l * {9}{c}}
        \toprule
        Dataset & \textbf{MMB} & \textbf{VQA\textsuperscript{v2}} & \textbf{GQA} & \textbf{MME} & \textbf{MM-Vet} & \textbf{VQA\textsuperscript{Text}} & \textbf{SQA\textsuperscript{IMG}} & \textbf{POPE} & \textbf{Avg.} \\
        \midrule
        \# Tokens & 58.3 & 73.5 & 81.3 & 69.3 & 65.2 & 83.4 & 48.3 & 80.1 & 74.1 \\
        \bottomrule
    \end{tabular}
    }
    \vspace{-0.2cm}
    \caption{\textbf{Adaptively decided token counts at $\theta=0.65$.}}
    \label{tab:token_per_benchmark}
    \vspace{-0.3cm}
\end{table}

\subsection{Discussion}

We further discuss a few other approaches that share common goal with ours.
Training-free token pruning~\cite{pdrop, chen2024fastv, hu2024illava, chen2025v2drop, ATP-LLaVA, lin2025vtw,vispruner, yang2025visionzip,prumerge,xu2024freepruner,divprune} reduces the number of visual tokens after feature extraction or at an intermediate LLM layer, typically by discarding or merging tokens based on attention scores or feature similarity.
These approaches offer similar advantage of being plug-and-play like ours, reducing inference cost without retraining.
Operating \textit{only} at inference time, however, they do not reduce training cost and often introduce a mismatch between the model's training dynamics and its pruned inference behavior, which can lead to substantial performance degradation under aggressive token reduction.
In contrast, our method generates a compact token set \textit{upfront}, guided by LLM supervision.
This allows the semantic structure to be learned end-to-end, maintaining full compatibility with the pre-trained LLM.

Chat-Univi~\cite{jin2024chat} and SeTok~\cite{setok} share a common goal with our method in generating semantically meaningful visual tokens, but differ in focus and design.
In particular, Chat-Univi generates a fixed number of tokens to represent images, where each patch contributes to multiple tokens, which contrasts with our objective of semantic disentanglement and adaptive token allocation.
SeTok treats tokenization as a part of the vision encoder, requiring \textit{large-scale supervised training} with object-level annotations and multiple transformer layers for token refinement.
In contrast, our approach serves as a lightweight projector trained end-to-end with the LLM, requiring only a single forward pass and no additional supervision.
While SeTok specializes in object-level representations, our method flexibly controls semantic granularity, enabling a broader token spectrum.

\section{Related Work}

\textbf{Multimodal Large Language Models.}
MLLMs extend text-only LLMs by incorporating vision encoders and lightweight adapters, enabling image understanding, visual reasoning, and multi-image or video based tasks~\cite{liu2023visual,li2023blip}.  
Early models typically pair LLMs with pre-trained vision encoders through simple projection~\cite{shikra, liu2023visual}, while recent systems introduce stronger vision backbones, improved alignment objectives, and richer multimodal instruction tuning~\cite{chen2024internvl,wang2024qwen2,lin2024vila}.
In parallel, compact MLLMs leverage smaller LLMs (\textit{e.g.}, Phi~\cite{abdin2024phi3}, Gemma~\cite{team2024gemma}, Minicpm~\cite{hu2024minicpm}) with improved adapters and data-efficient training~\cite{chu2024mobilevlm,beyer2024paligemma,zhu2024llavaphi,yao2024minicpmv}.
However, across these models, images are still converted to a long sequence of fixed patch-level tokens, leading to semantically entangled and redundant representations that inflate KV-cache size and latency without commensurate accuracy gains.
We resolve this long-standing bottleneck by selecting semantically aligned tokens with an adaptively sized budget tied to image complexity.

\vspace{0.1cm} \noindent
\textbf{Efficient MLLMs.}
On the language side, simplified fusion~\cite{gu2024mamba} and inference accelerations such as layer skipping or speculative/adaptive decoding~\cite{elhoushi2024layerskip,leviathan2023speculative,zhu2024adaptivedec} have been applied to reduce parameters or computational overhead.
On the other hand, projector-centric methods compress vision features prior to the LLM via query-based summarization~\cite{li2023blip, dai2023instructblip}, convolution layers that reduce spatial resolution~\cite{cha2024honeybee, chu2024mobilevlm}, and grid-wise aggregation that downsamples tokens via spatial grouping or patch aggregation~\cite{pixel-shuffle, li2025tokenpacker}.
While effective for bandwidth, these techniques largely view the projector as a feature compressor that preserves proximity rather than semantics;
the resulting tokens often remain semantically entangled.
In contrast, we form semantically coherent visual words, and align them with the LLM's discrete interface, boosting reasoning at equal or lower token budgets without encoder-specific heads. 

Another major paradigm focuses on eliminating redundant tokens at inference.
Intra-LLM pruning~\cite{pdrop, chen2024fastv, hu2024illava, chen2025v2drop, ATP-LLaVA, lin2025vtw,vispruner} discards tokens at intermediate layers based on criteria such as layer index, attention-score, or with lightweight pruning modules. 
However, these approaches operate post-hoc on already entangled patch tokens and can interact unpredictably with kernel-level optimizations like KV-cache policies or FlashAttention.
To mitigate this, a pre-LLM selection strategy~\cite{yang2025visionzip,prumerge,xu2024freepruner,divprune} is adopted to filter visual tokens before entering the language model, commonly by measuring their similarity to a global CLS token or text embedding, or by promoting token diversity to preserve complementary visual cues.
These methods improve efficiency by reducing token budgets prior to multimodal fusion, but they primarily perform compression rather than semantic alignment.
Instead, we address the root cause by constructing semantically aligned tokens upfront.

Most aligned with our work, tokenizers grouping patches into variable-length semantic units~\cite{setok,slotmllm,jin2024chat} have been proposed recently.
While effective for grounding, they often require auxiliary supervision, reconstruction losses, or multi-stage refinement, increasing training cost and reducing generality.
Moreover, token granularity is frequently controlled only indirectly, and adjusting budgets can require retraining or some heuristic.
Our method targets the same goal but with a lighter, more general mechanism: feature-space clustering trained solely by the LLM's objective.
A single similarity threshold controls granularity and an adaptive token budget, tunable at inference without retraining.


\section{Summary}
\label{sec:conclusion}
Motivated by mismatched properties between the visual and text tokens, we introduce a novel visual projector in MLLMs to address visual-text modality gap.
Our method produces compact visual tokens in which each token corresponds to a semantically coherent concept, effectively mitigating the entanglement, redundancy, and loss of details inherent in conventional projectors.
By dynamically allocating token counts based on image complexity and controlling semantic granularity through a single similarity threshold, DiVT adapts naturally to diverse visual inputs while supporting training-free adjustment at inference.
Across benchmarks, this semantically aligned representation achieves competitive or superior performance with only a small fraction of the original tokens, substantially reducing memory and latency.

\clearpage
\section*{Acknowledgments}
This work was also supported by the SOFT Foundry Institute at SNU, Samsung Electronics, Youlchon Foundation, National Research Foundation of Korea (NRF) grants (RS-2021-NR05515, RS-2024-00336576, RS-2023-0022663, RS-2025-25399604, RS-2024-00342044, RS-2025-16063688, RS-2025-02215813), and the Institute for Information \& Communication Technology Planning \& Evaluation (IITP) grants (RS-2022-II220264, RS-2024-00353131) funded by the Korean government.

{
    \small
    \bibliographystyle{ieeenat_fullname}
    \bibliography{main}

@String(CVPR= {IEEE Conf. Comput. Vis. Pattern Recog.})

@String(ICCV= {Int. Conf. Comput. Vis.})

@String(ECCV= {Eur. Conf. Comput. Vis.})

@String(ICLR = {Int. Conf. Learn. Represent.})

@String(AAAI = {AAAI})

@String(CVPR  = {CVPR})

@String(ICCV  = {ICCV})

@String(ECCV  = {ECCV})

@String(NeurIPS  = {NeurIPS})

@String(ICML  = {ICML})

@String(ICLR  = {ICLR})

@String(ACL = {ACL})

@String(EMNLP = {EMNLP})

@inproceedings{divprune,
  title={Div{P}rune: Diversity-based visual token pruning for large multimodal models},
  author={Alvar, Saeed Ranjbar and Singh, Gursimran and Akbari, Mohammad and Zhang, Yong},
  booktitle=CVPR,
  year={2025}
}

@article{slotmllm,
  title={Slot-{MLLM}: Object-Centric Visual Tokenization for Multimodal {LLM}},
  author={Chi, Donghwan and Kim, Hyomin and Oh, Yoonjin and Kim, Yongjin and Lee, Donghoon and Jo, Daejin and Kim, Jongmin and Baek, Junyeob and Ahn, Sungjin and Kim, Sungwoong},
  journal={arXiv:2505.17726},
  year={2025}
}

@article{setok,
  title={Towards semantic equivalence of tokenization in multimodal llm},
  author={Wu, Shengqiong and Fei, Hao and Li, Xiangtai and Ji, Jiayi and Zhang, Hanwang and Chua, Tat-Seng and Yan, Shuicheng},
  journal={arXiv:2406.05127},
  year={2024}
}

@inproceedings{liu2023visual,
  title={Visual instruction tuning},
  author={Liu, Haotian and Li, Chunyuan and Wu, Qingyang and Lee, Yong Jae},
  booktitle=NeurIPS,
  year={2023},
}

@inproceedings{li2023blip,
  title={{BLIP}-2: Bootstrapping language-image pre-training with frozen image encoders and large language models},
  author={Li, Junnan and Li, Dongxu and Savarese, Silvio and Hoi, Steven},
  booktitle=ICML,
  year={2023}
}

@inproceedings{chen2024internvl,
  title={Intern{VL}: Scaling up vision foundation models and aligning for generic visual-linguistic tasks},
  author={Chen, Zhe and Wu, Jiannan and Wang, Wenhai and Su, Weijie and Chen, Guo and Xing, Sen and Zhong, Muyan and Zhang, Qinglong and Zhu, Xizhou and Lu, Lewei and others},
  booktitle=CVPR,
  year={2024}
}

@article{wang2024qwen2,
  title={Qwen2-{VL}: Enhancing vision-language model's perception of the world at any resolution},
  author={Wang, Peng and Bai, Shuai and Tan, Sinan and Wang, Shijie and Fan, Zhihao and Bai, Jinze and Chen, Keqin and Liu, Xuejing and Wang, Jialin and Ge, Wenbin and others},
  journal={arXiv:2409.12191},
  year={2024}
}

@inproceedings{lin2024vila,
  title={{VILA}: On pre-training for visual language models},
  author={Lin, Ji and Yin, Hongxu and Ping, Wei and Molchanov, Pavlo and Shoeybi, Mohammad and Han, Song},
  booktitle=CVPR,
  year={2024}
}

@inproceedings{dai2023instructblip,
  title={Instruct{BLIP}: Towards general-purpose vision-language models with instruction tuning},
  author={Dai, Wenliang and Li, Junnan and Li, Dongxu and Tiong, Anthony and Zhao, Junqi and Wang, Weisheng and Li, Boyang and Fung, Pascale N and Hoi, Steven},
  booktitle=NeurIPS,
  year={2023}
}

@article{chu2024mobilevlm,
  title={Mobile{VLM} V2: Faster and Stronger Baseline for Vision Language Model},
  author={Chu, Xiangxiang and Qiao, Limeng and Zhang, Xinyu and Xu, Shuang and Wei, Fei and Yang, Yang and Sun, Xiaofei and Hu, Yiming and Lin, Xinyang and Zhang, Bo and others},
  journal={arXiv:2402.03766},
  year={2024}
}

@article{beyer2024paligemma,
  title={Pali{G}emma: A versatile 3{B} {VLM} for transfer},
  author={Beyer, Lucas and Steiner, Andreas and Pinto, Andr{\'e} Susano and Kolesnikov, Alexander and Wang, Xiao and Salz, Daniel and Neumann, Maxim and Alabdulmohsin, Ibrahim and Tschannen, Michael and Bugliarello, Emanuele and others},
  journal={arXiv:2407.07726},
  year={2024}
}

@article{li2025tokenpacker,
  title={Tokenpacker: Efficient visual projector for multimodal {LLM}},
  author={Li, Wentong and Yuan, Yuqian and Liu, Jian and Tang, Dongqi and Wang, Song and Qin, Jie and Zhu, Jianke and Zhang, Lei},
  journal={International Journal of Computer Vision},
  pages={1--19},
  year={2025},
  publisher={Springer}
}

@inproceedings{prumerge,
  title={{LLaVA-PruMerge}: Adaptive token reduction for efficient large multimodal models},
  author={Shang, Yuzhang and Cai, Mu and Xu, Bingxin and Lee, Yong Jae and Yan, Yan},
  booktitle=ICCV,
  year={2025}
}

@article{chen2025v2drop,
  title={Variation-aware Vision Token Dropping for Faster Large Vision-Language Models},
  author={Chen, Junjie and Liu, Xuyang and Wen, Zichen and Wang, Yiyu and Huang, Siteng and Chen, Honggang},
  journal={arXiv:2509.01552},
  year={2025}
}

@article{abdin2024phi3,
  title={Phi-3 Technical Report: A Highly Capable Language Model Locally on Your Phone},
  author={Abdin, Marah and Aneja, Jyoti and Awadalla, Hany and Awadallah, Ahmed and Awan, Ammar Ahmad and Bach, Nguyen and Bahree, Amit and Bakhtiari, Arash and Bao, Jianmin and Behl, Harkirat and others},
  journal={arXiv:2404.14219},
  year={2024}
}

@article{hu2024minicpm,
  title={Minicpm: Unveiling the potential of small language models with scalable training strategies},
  author={Hu, Shengding and Tu, Yuge and Han, Xu and He, Chaoqun and Cui, Ganqu and Long, Xiang and Zheng, Zhi and Fang, Yewei and Huang, Yuxiang and Zhao, Weilin and others},
  journal={arXiv preprint arXiv:2404.06395},
  year={2024}
}

@article{yao2024minicpmv,
  title={{MiniCPM-V}: A {GPT-4V} level mllm on your phone},
  author={Yao, Yuan and Yu, Tianyu and Zhang, Ao and Wang, Chongyi and Cui, Junbo and Zhu, Hongji and Cai, Tianchi and Li, Haoyu and Zhao, Weilin and He, Zhihui and others},
  journal={arXiv:2408.01800},
  year={2024}
}

@article{team2024gemma,
  title={Gemma: Open models based on gemini research and technology},
  author={Team, Gemma and Mesnard, Thomas and Hardin, Cassidy and Dadashi, Robert and Bhupatiraju, Surya and Pathak, Shreya and Sifre, Laurent and Rivi{\`e}re, Morgane and Kale, Mihir Sanjay and Love, Juliette and others},
  journal={arXiv:2403.08295},
  year={2024}
}

@inproceedings{zhu2024llavaphi,
  title={{LLaVA-Phi}: Efficient multi-modal assistant with small language model},
  author={Zhu, Yichen and Zhu, Minjie and Liu, Ning and Xu, Zhiyuan and Peng, Yaxin},
  booktitle={International Workshop on Efficient Multimedia Computing under Limited},
  year={2024}
}

@inproceedings{cha2024honeybee,
  title={Honeybee: Locality-enhanced projector for multimodal {LLM}},
  author={Cha, Junbum and Kang, Wooyoung and Mun, Jonghwan and Roh, Byungseok},
  booktitle=CVPR,
  year={2024}
}

@inproceedings{gu2024mamba,
  title={Mamba: Linear-time sequence modeling with selective state spaces},
  author={Gu, Albert and Dao, Tri},
  booktitle={Conference on language modeling},
  year={2024}
}

@inproceedings{leviathan2023speculative,
  title={Fast inference from transformers via speculative decoding},
  author={Leviathan, Yaniv and Kalman, Matan and Matias, Yossi},
  booktitle=ICML,
  year={2023}
}

@inproceedings{zhu2024adaptivedec,
title={Improving Open-Ended Text Generation via Adaptive Decoding},
author={Wenhong Zhu and Hongkun Hao and Zhiwei He and Yiming Ai and Rui Wang},
booktitle=ICML,
year={2024},
}

@inproceedings{elhoushi2024layerskip,
  title={Layerskip: Enabling early exit inference and self-speculative decoding},
  author={Elhoushi, Mostafa and Shrivastava, Akshat and Liskovich, Diana and Hosmer, Basil and Wasti, Bram and Lai, Liangzhen and Mahmoud, Anas and Acun, Bilge and Agarwal, Saurabh and Roman, Ahmed and others},
  booktitle=ACL,
  year={2024}
}

@INPROCEEDINGS{pdrop,
  author={Xing, Long and Huang, Qidong and Dong, Xiaoyi and Lu, Jiajie and Zhang, Pan and Zang, Yuhang and Cao, Yuhang and He, Conghui and Wang, Jiaqi and Wu, Feng and Lin, Dahua},
  booktitle=CVPR, 
  title={Conical Visual Concentration for Efficient Large Vision-Language Models}, 
  year={2025}
}

@inproceedings{chen2024fastv,
  title={An image is worth 1/2 tokens after layer 2: Plug-and-play inference acceleration for large vision-language models},
  author={Chen, Liang and Zhao, Haozhe and Liu, Tianyu and Bai, Shuai and Lin, Junyang and Zhou, Chang and Chang, Baobao},
  booktitle=ECCV,
  year={2024}
}

@article{hu2024illava,
  title={{iLLaVA}: An image is worth fewer than 1/3 input tokens in large multimodal models},
  author={Hu, Lianyu and Shang, Fanhua and Wan, Liang and Feng, Wei},
  journal={arXiv:2412.06263},
  year={2024}
}

@inproceedings{ATP-LLaVA,
  title={{ATP}-{LLaVA}: Adaptive token pruning for large vision language models},
  author={Ye, Xubing and Gan, Yukang and Ge, Yixiao and Zhang, Xiao-Ping and Tang, Yansong},
  booktitle=CVPR,
  year={2025}
}

@inproceedings{lin2025vtw,
  title={Boosting multimodal large language models with visual tokens withdrawal for rapid inference},
  author={Lin, Zhihang and Lin, Mingbao and Lin, Luxi and Ji, Rongrong},
  booktitle=AAAI,
  year={2025}
}

@inproceedings{yang2025visionzip,
  title={Vision{Z}ip: Longer is better but not necessary in vision language models},
  author={Yang, Senqiao and Chen, Yukang and Tian, Zhuotao and Wang, Chengyao and Li, Jingyao and Yu, Bei and Jia, Jiaya},
  booktitle=CVPR,
  year={2025}
}

@article{xu2024freepruner,
  title={free{P}runer: A training-free approach for large multimodal model acceleration},
  author={Xu, Bingxin and Shang, Yuzhang and Ge, Yunhao and Lou, Qian and Yan, Yan},
  journal={arXiv:2411.15446},
  year={2024}
}

@inproceedings{vispruner,
  title={Beyond text-visual attention: Exploiting visual cues for effective token pruning in vlms},
  author={Zhang, Qizhe and Cheng, Aosong and Lu, Ming and Zhang, Renrui and Zhuo, Zhiyong and Cao, Jiajun and Guo, Shaobo and She, Qi and Zhang, Shanghang},
  booktitle=ICCV,
  year={2025}
}

@inproceedings{clip,
  title={Learning transferable visual models from natural language supervision},
  author={Radford, Alec and Kim, Jong Wook and Hallacy, Chris and Ramesh, Aditya and Goh, Gabriel and Agarwal, Sandhini and Sastry, Girish and Askell, Amanda and Mishkin, Pamela and Clark, Jack and others},
  booktitle=ICML,
  year={2021}
}

@inproceedings{SigLIP,
  title={Sigmoid loss for language image pre-training},
  author={Zhai, Xiaohua and Mustafa, Basil and Kolesnikov, Alexander and Beyer, Lucas},
  booktitle=ICCV,
  year={2023}
}

@article{Dinov2,
  title={{DINO}v2: Learning robust visual features without supervision},
  author={Oquab, Maxime and Darcet, Timoth{\'e}e and Moutakanni, Th{\'e}o and Vo, Huy and Szafraniec, Marc and Khalidov, Vasil and Fernandez, Pierre and Haziza, Daniel and Massa, Francisco and El-Nouby, Alaaeldin and others},
  journal={arXiv:2304.07193},
  year={2023}
}

@inproceedings{vit,
title={An Image is Worth 16x16 Words: Transformers for Image Recognition at Scale},
author={Alexey Dosovitskiy and Lucas Beyer and Alexander Kolesnikov and Dirk Weissenborn and Xiaohua Zhai and Thomas Unterthiner and Mostafa Dehghani and Matthias Minderer and Georg Heigold and Sylvain Gelly and Jakob Uszkoreit and Neil Houlsby},
booktitle=ICLR,
year={2021}
}

@inproceedings{ToME,
title={Token Merging: Your {ViT} But Faster},
author={Daniel Bolya and Cheng-Yang Fu and Xiaoliang Dai and Peizhao Zhang and Christoph Feichtenhofer and Judy Hoffman},
booktitle=ICLR,
year={2023}
}

@article{shikra,
  title={Shikra: Unleashing multimodal {LLM}'s referential dialogue magic},
  author={Chen, Keqin and Zhang, Zhao and Zeng, Weili and Zhang, Richong and Zhu, Feng and Zhao, Rui},
  journal={arXiv:2306.15195},
  year={2023}
}

@article{qwenvl,
  title={Qwen technical report},
  author={Bai, Jinze and Bai, Shuai and Chu, Yunfei and Cui, Zeyu and Dang, Kai and Deng, Xiaodong and Fan, Yang and Ge, Wenbin and Han, Yu and Huang, Fei and others},
  journal={arXiv:2309.16609},
  year={2023}
}

@inproceedings{mmbench,
  title={{MMBench}: Is your multi-modal model an all-around player?},
  author={Liu, Yuan and Duan, Haodong and Zhang, Yuanhan and Li, Bo and Zhang, Songyang and Zhao, Wangbo and Yuan, Yike and Wang, Jiaqi and He, Conghui and Liu, Ziwei and others},
  booktitle=ECCV,
  year={2024}
}

@inproceedings{vqav2,
  title={Making the {V} in {VQA} matter: Elevating the role of image understanding in visual question answering},
  author={Goyal, Yash and Khot, Tejas and Summers-Stay, Douglas and Batra, Dhruv and Parikh, Devi},
  booktitle=CVPR,
  year={2017}
}

@inproceedings{gqa,
  title={{GQA}: A new dataset for real-world visual reasoning and compositional question answering},
  author={Hudson, Drew A and Manning, Christopher D},
  booktitle=CVPR,
  year={2019}
}

@inproceedings{textvqa,
  title={Towards vqa models that can read},
  author={Singh, Amanpreet and Natarajan, Vivek and Shah, Meet and Jiang, Yu and Chen, Xinlei and Batra, Dhruv and Parikh, Devi and Rohrbach, Marcus},
  booktitle=CVPR,
  year={2019}
}

@inproceedings{pope,
    title = "Evaluating Object Hallucination in Large Vision-Language Models",
    author = "Li, Yifan  and
      Du, Yifan  and
      Zhou, Kun  and
      Wang, Jinpeng  and
      Zhao, Xin  and
      Wen, Ji-Rong",
    booktitle = EMNLP,
    year = "2023"
}

@inproceedings{mm-vet,
    author = {Yu, Weihao and Yang, Zhengyuan and Li, Linjie and Wang, Jianfeng and Lin, Kevin and Liu, Zicheng and Wang, Xinchao and Wang, Lijuan},
    title = {{MM-Vet}: evaluating large multimodal models for integrated capabilities},
    year = {2024},
    booktitle = ICML,
}

@inproceedings{sqa-img,
  title={Learn to explain: Multimodal reasoning via thought chains for science question answering},
  author={Lu, Pan and Mishra, Swaroop and Xia, Tanglin and Qiu, Liang and Chang, Kai-Wei and Zhu, Song-Chun and Tafjord, Oyvind and Clark, Peter and Kalyan, Ashwin},
  booktitle=NeurIPS,
  year={2022}
}

@article{mme,
  title={Mme: A comprehensive evaluation benchmark for multimodal large language models},
  author={Fu, Chaoyou and Chen, Peixian and Shen, Yunhang and Qin, Yulei and Zhang, Mengdan and Lin, Xu and Yang, Jinrui and Zheng, Xiawu and Li, Ke and Sun, Xing and others},
  journal={arXiv preprint arXiv:2306.13394},
  year={2023}
}

@article{pixel-shuffle,
  title={How far are we to {GPT}-4{V}? closing the gap to commercial multimodal models with open-source suites},
  author={Chen, Zhe and Wang, Weiyun and Tian, Hao and Ye, Shenglong and Gao, Zhangwei and Cui, Erfei and Tong, Wenwen and Hu, Kongzhi and Luo, Jiapeng and Ma, Zheng and others},
  journal={Science China Information Sciences},
  volume={67},
  number={12},
  pages={220101},
  year={2024},
  publisher={Springer}
}

@inproceedings{flashsloth,
  title={Flash{S}loth: Lightning multimodal large language models via embedded visual compression},
  author={Tong, Bo and Lai, Bokai and Zhou, Yiyi and Luo, Gen and Shen, Yunhang and Li, Ke and Sun, Xiaoshuai and Ji, Rongrong},
  booktitle=CVPR,
  year={2025}
}

@inproceedings{bytepairencoding,
  title={Neural machine translation of rare words with subword units},
  author={Sennrich, Rico and Haddow, Barry and Birch, Alexandra},
  booktitle=ACL,
  year={2016}
}

@inproceedings{jin2024chat,
  title={Chat-univi: Unified visual representation empowers large language models with image and video understanding},
  author={Jin, Peng and Takanobu, Ryuichi and Zhang, Wancai and Cao, Xiaochun and Yuan, Li},
  booktitle=CVPR,
  year={2024}
}

@misc{vicuna2023,
    title = {Vicuna: An Open-Source Chatbot Impressing GPT-4 with 90\%* ChatGPT Quality},
    url = {https://lmsys.org/blog/2023-03-30-vicuna/},
    author = {Chiang, Wei-Lin and Li, Zhuohan and Lin, Zi and Sheng, Ying and Wu, Zhanghao and Zhang, Hao and Zheng, Lianmin and Zhuang, Siyuan and Zhuang, Yonghao and Gonzalez, Joseph E. and Stoica, Ion and Xing, Eric P.},
    month = {March},
    year = {2023}
}
}

\clearpage
\setcounter{page}{1}
\maketitlesupplementary

\appendix

\pagenumbering{roman}
\renewcommand\thetable{\Roman{table}}
\renewcommand\thefigure{\Roman{figure}}
\setcounter{page}{1}
\setcounter{section}{0}
\setcounter{table}{0}
\setcounter{figure}{0}

\section{Implementation Details}
\label{appendix:implementation}

\textbf{Hyperparameters.}
Our implementation closely follows the standard LLaVA-1.5 training pipeline, with only minimal adjustments required to integrate \method\ into the multimodal architecture.
Pretraining is performed for one epoch following the LLaVA-1.5 recipe; we use batch size of 256, initial learning rate of $10^{-3}$ with cosine decay and 3\% warm-up, AdamW without weight decay, and DeepSpeed Stage~2.
We update only the parameters of our projector.
Finetuning adopts the same training configuration, differing only in the decreased batch size of 128, a reduced learning rate of $2\!\times\!10^{-5}$, and DeepSpeed Stage~3, through which the projector and LLM are jointly optimized.
Unless stated otherwise, we adopt $\theta = 0.65$ as the default threshold for our \method.

\vspace{0.1cm} \noindent
\textbf{More on Model Architecture.}
We use two ViT-based vision encoders, \texttt{facebook/dinov2-large} and \texttt{google/siglip-large-patch16-384} for experiment in \cref{tab:backbone_comparison}.
Since DINOv2 is originally trained with $224 \times 224$ inputs and therefore outputs merely 256 patch embeddings, we modify its preprocessing to accept $336 \times 336$ resolution so that it produces 576 patch features, making its output shape consistent with CLIP or SigLIP encoders.
All experiments are conducted on a compute cluster equipped with eight NVIDIA RTX A6000 GPUs (48GB).

\section{Performance with Various Thresholds $\theta$}
\label{appendix:threshold}

A key advantage of \method\ is that the similarity threshold $\theta$ controls semantic granularity of the clusters.
We experiment with $\theta \in \{0.3, 0.4, 0.5, 0.62, 0.65, 0.75, 0.8\}$, spanning coarse to highly fine-grained clustering regimes.

\cref{tab:token_budget_analysis} provides the full numerical results together with the corresponding average token counts.
The table clearly shows a larger $\theta$ expands the token budget and this increase tends to align with the observed performance gains across most benchmarks.
Once the threshold $\theta$ reaches beyond $0.75$, however, the clusters become overly fragmented and semantically redundant, leading to a mild degradation in accuracy despite further increases in token count.
This behavior confirms that a moderate granularity provides the best trade-off between accuracy and token efficiency.
\begin{table*}
    \small
    \centering
    \resizebox{0.8\textwidth}{!}{%
        \begin{tabular}{l * {9}{c}}
            \toprule
            \textbf{$\theta$} & \textbf{\# Tokens} & \textbf{MMB} & \textbf{VQA\textsuperscript{v2}} & \textbf{GQA} & \textbf{MME} & \textbf{MM-Vet} & \textbf{VQA\textsuperscript{Text}} & \textbf{SQA\textsuperscript{IMG}} & \textbf{POPE} \\
            \midrule
            0.3  & 13.5  & 64.2 & 75.3 & 59.2 & 1462.8 & 28.0 & 55.4 & 69.4 & 84.3 \\
            0.4  & 22.4  & 64.7 & 76.4 & 60.1 & 1450.9 & 31.7 & 56.1 & 69.1 & 84.8 \\
            0.5  & 35.7  & 65.0 & 77.0 & 60.6 & 1458.2 & 31.7 & 57.1 & 68.2 & 85.8 \\
            0.62 & 63.7  & 64.3 & 77.7 & 61.6 & 1463.0 & 30.6 & 57.0 & \textbf{70.6} & \textbf{86.2} \\
            0.65 & 74.1  & 65.5 & 77.7 & 61.4 & \textbf{1465.7} & \textbf{32.1} & 57.2 & 68.1 & 85.8 \\
            0.75 & 136.5 & \textbf{66.7} & \textbf{78.2} & \textbf{62.0} & 1457.6 & 30.2 & \textbf{57.7} & 70.0 & \textbf{86.2} \\
            0.8  & 175.3 & 65.3 & \textbf{78.2} & 61.9 & 1456.5 & 31.3 & 57.4 & 68.8 & 85.8 \\
            \bottomrule
        \end{tabular}
    }
    \caption{\textbf{Performance of our DiVT under varying similarity thresholds}}
    \label{tab:token_budget_analysis}
\end{table*}

\section{Training and Inference Time Analysis}
\label{appendix:appendix_efficiency}

\cref{tab:efficiency} summarizes the computational advantages of \method.
By treating substantially fewer visual tokens than the 576-token MLP baseline, our \method\ significantly shortens the multimodal forward pass and leads to notable speedups at both training and inference.

\begin{table*}[h]
    \small
    \centering
    \begin{tabular}{lccccc}
        \toprule
        \textbf{Method} &
        \textbf{Pretraining (h)} &
        \textbf{Finetuning (h)} &
        \textbf{Inference (h)} &
        \textbf{KV-Cache (MB)} &
        \textbf{Prefill Latency (ms)} \\
        \midrule
        MLP Projector        
            & 5.7 {\scriptsize\textcolor{gray}{(100\%)}} 
            & 20.0 {\scriptsize\textcolor{gray}{(100\%)}} 
            & 5.5 {\scriptsize\textcolor{gray}{(100\%)}} 
            & 288.0 {\scriptsize\textcolor{gray}{(100\%)}} 
            & 138.2 {\scriptsize\textcolor{gray}{(100\%)}} \\
        \method$_{\theta=0.4}$  
            & 1.1 {\scriptsize\textcolor{gray}{(19.3\%)}} 
            & 12.9 {\scriptsize\textcolor{gray}{(64.5\%)}} 
            & 3.4 {\scriptsize\textcolor{gray}{(61.8\%)}} 
            & 11.0 {\scriptsize\textcolor{gray}{(3.8\%)}} 
            & 71.3 {\scriptsize\textcolor{gray}{(51.6\%)}} \\
        \method$_{\theta=0.5}$  
            & 1.4 {\scriptsize\textcolor{gray}{(24.6\%)}} 
            & 13.1 {\scriptsize\textcolor{gray}{(65.5\%)}} 
            & 3.6 {\scriptsize\textcolor{gray}{(65.5\%)}} 
            & 17.6 {\scriptsize\textcolor{gray}{(6.1\%)}} 
            & 76.6 {\scriptsize\textcolor{gray}{(55.4\%)}} \\
        \method$_{\theta=0.65}$ 
            & 1.9 {\scriptsize\textcolor{gray}{(33.3\%)}} 
            & 13.7 {\scriptsize\textcolor{gray}{(68.5\%)}} 
            & 3.9 {\scriptsize\textcolor{gray}{(70.9\%)}} 
            & 36.8 {\scriptsize\textcolor{gray}{(12.8\%)}} 
            & 104.4 {\scriptsize\textcolor{gray}{(75.6\%)}} \\
        \method$_{\theta=0.75}$ 
            & 2.7 {\scriptsize\textcolor{gray}{(47.4\%)}} 
            & 14.5 {\scriptsize\textcolor{gray}{(72.5\%)}} 
            & 4.7 {\scriptsize\textcolor{gray}{(85.5\%)}} 
            & 68.1 {\scriptsize\textcolor{gray}{(23.6\%)}} 
            & 138.3 {\scriptsize\textcolor{gray}{(100.1\%)}} \\
        \bottomrule
    \end{tabular}
    \caption{
    \textbf{Training and inference cost of \method\ across different similarity thresholds}.
    Training time is measured using eight RTX~A6000 GPUs, and inference time is measured on the VQAv2 evaluation set using a single RTX~A6000 GPU.
    KV-cache memory is computed analytically from the LLaMA-7B architecture, where each visual token contributes approximately 0.5\,MB of KV-cache.
    Prefill latency is measured by averaging multiple stable forward passes after warm-up.
    }
    \label{tab:efficiency}
\end{table*}
The efficiency comparison in \cref{tab:efficiency} highlights how \method\ substantially reduces computational cost across pretraining, finetuning, inference, and KV-cache memory usage.
The largest improvement appears in the pretraining stage, where the LLM is frozen and the computational cost is largely determined by the sequence length processed through each transformer layer. 
Lower thresholds significantly shorten this sequence, leading to proportionally large reductions in attention computation and, in turn, overall pretraining time.
Finetuning likewise benefits from the reduced token count, though the gains are somewhat moderated by the need to update the full LLM. 
Still, the lighter visual sequence consistently improves optimization efficiency, providing meaningful savings in both training phases.

Inference reveals additional practical advantages.
Because the sequence length linearly scales KV-cache memory, reducing the number of tokens shrinks the KV-cache footprint by over 90\% at coarse thresholds such as $\theta{=}0.4$.
Such reductions substantially ease the memory burden and suggest potential scalability benefits for scenarios involving multi-images or video, where limited KV-cache capacity and context length frequently become bottlenecks. 

Prefill latency is influenced both by the number of visual tokens and by the cost of our clustering algorithm. 
At low thresholds, the substantial reduction in token count dominates the overall computation, making the clustering overhead comparatively minor and enabling nearly a 2× speedup over the MLP projector. 
Even at $\theta{=}0.75$, where the clustering step becomes slightly heavier, the resulting prefill latency remains close to that of the MLP baseline, indicating that the additional cost introduced by clustering is not a major bottleneck in practice.
Crucially, this overhead appears only once during the prefill stage. 
After tokenization, the subsequent decoding process depends solely on the final number of visual tokens, not on how they were formed. 
As a result, \method\ benefits from reduced inference-time computation throughout the entire generation process, whereas the MLP projector must continue to handle a much longer visual sequence at every decoding step. 
This separation demonstrates that the overhead associated with clustering is limited while the gains in end-to-end efficiency are substantial.

Overall, the threshold parameter $\theta$ allows practitioners to modulate computational cost with a single control knob, ranging from highly compact and efficient configurations to more detailed representations when resources permit. 
This simple controllability, together with consistently lower KV-cache usage and reduced decoding cost, makes \method\ an appealing alternative to the MLP projector from an efficiency standpoint.

\section{Additional Attention Map Visualizations}
\label{appendix:appendix_attention}

\begin{figure*}
    \centering
    \includegraphics[width=0.95\linewidth]{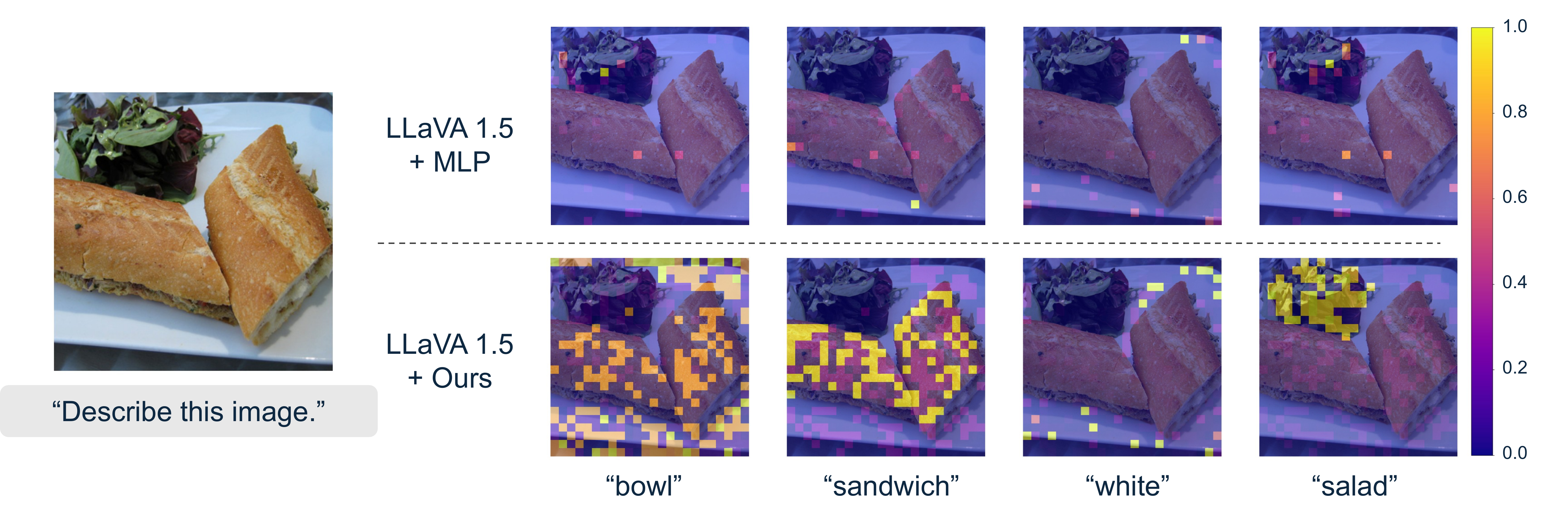}
    \includegraphics[width=0.95\linewidth]{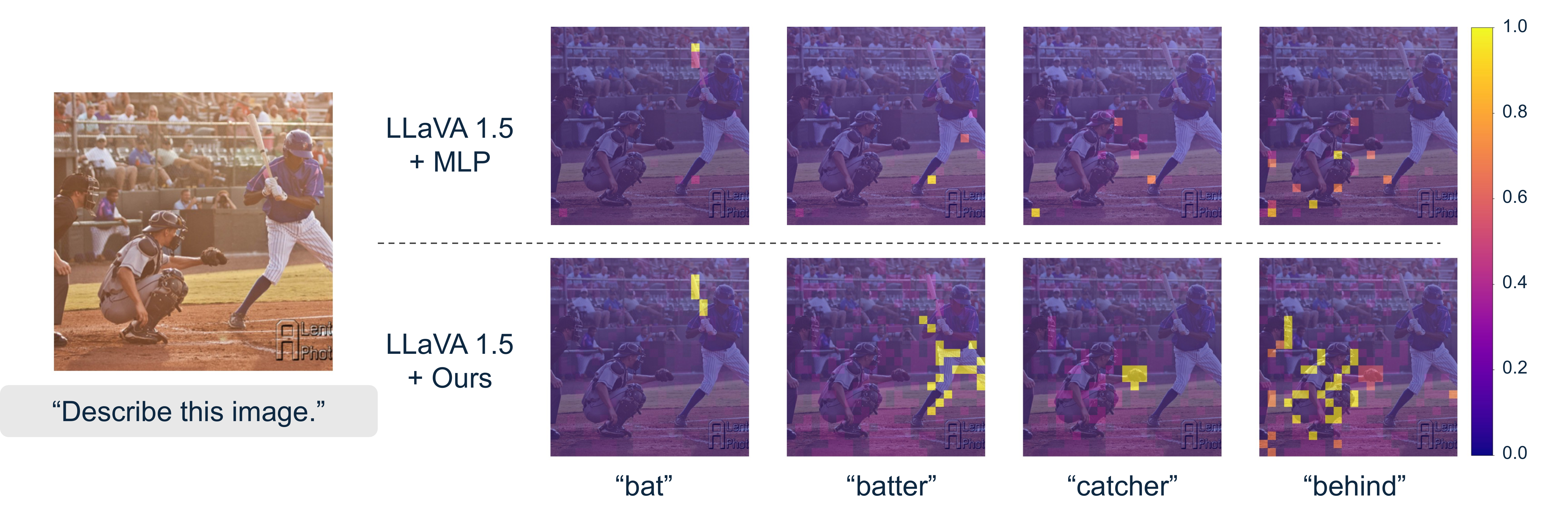}
    \includegraphics[width=0.95\linewidth]{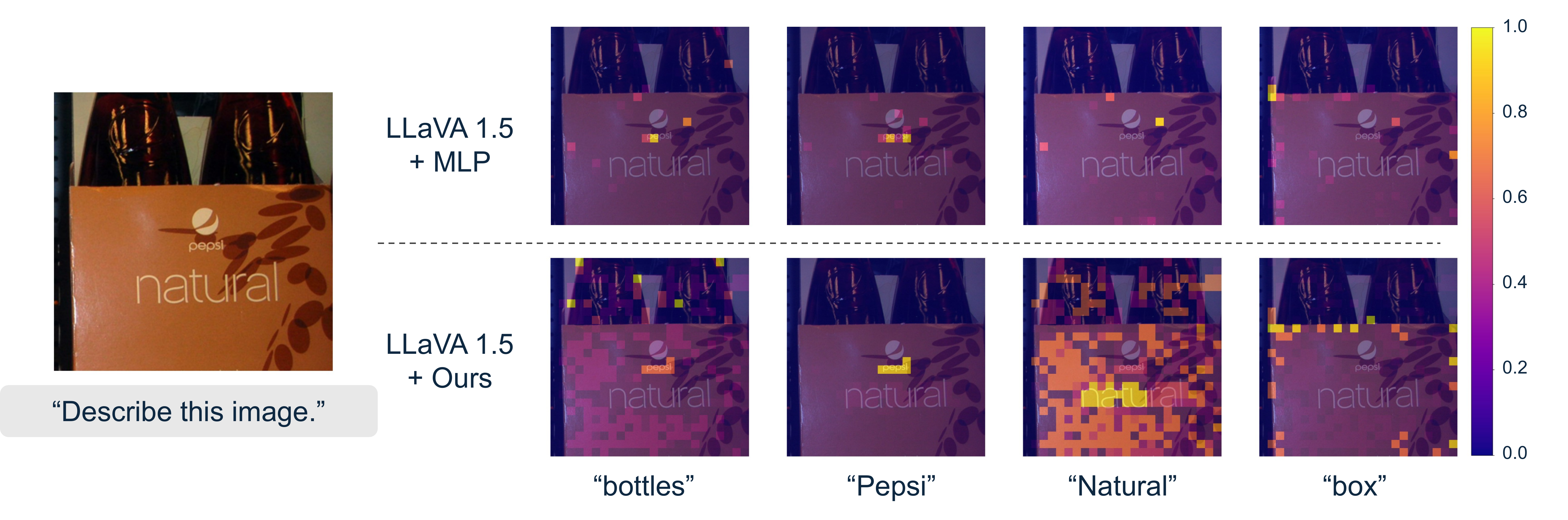}
    \includegraphics[width=0.95\linewidth]{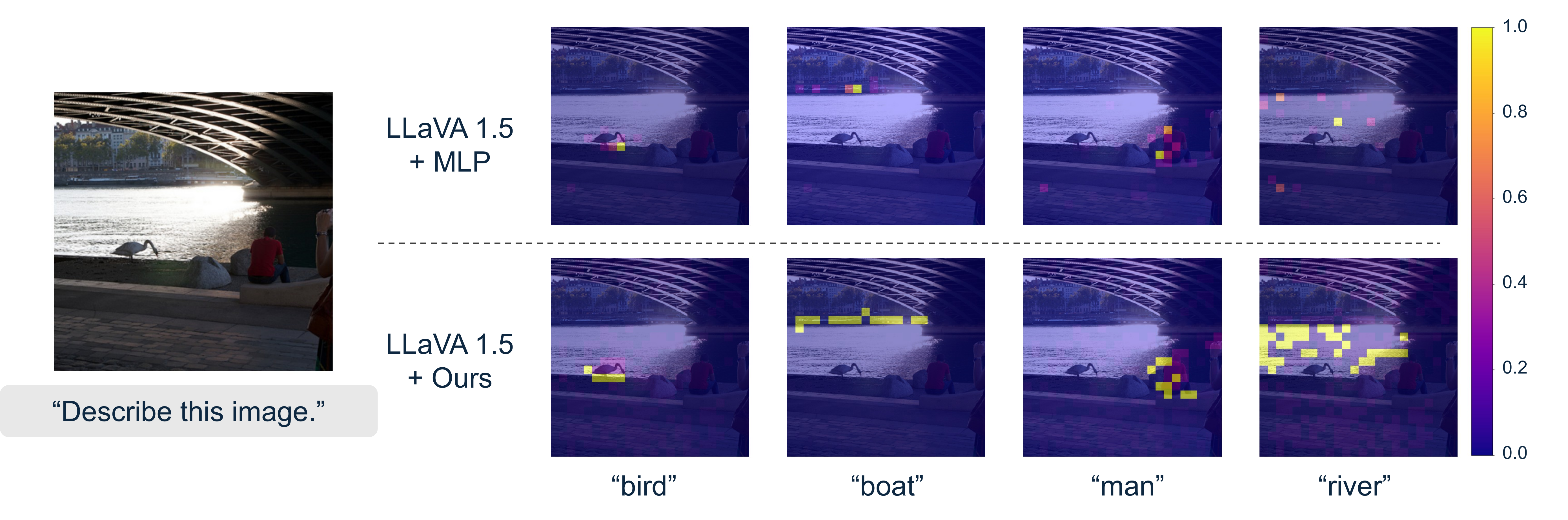}
    \caption{\textbf{Additional attention map comparisons between DiVT and the MLP projector}. The cluster-based representation in DiVT leads to more consistent and interpretable attention behavior across textual tokens.}
    \label{fig:extra_attention}
\end{figure*}

\cref{fig:extra_attention} illustrates additional examples of attention patterns comparing \method\ with the standard MLP projector.
To visualize the attention received by each textual token, we aggregate the attention weights assigned to a given \method\ token and project them onto all patches belonging to that token's cluster.
This cluster-level visualization highlights which semantic region the model relies on when it processes each text token.

Since \method\ (bottom) aggregates patches into coherent semantic clusters, the resulting attention maps reveal clear and localized patterns.
Each textual token tends to focus on a distinct visual concept, making the grounding behavior easy to interpret.
In contrast, MLP projectors (top) operate at the patch-level and thus often assign disproportionately high attention to a small subset of tokens, regardless of the query.
This obscures which visual evidence the model is using, thereby leading to diffused or noisy activation patterns and hurting interpretability.

\section{Additional Cluster Visualization}
\label{appendix:cluster_vis}

\begin{figure*}
    \centering
    \includegraphics[width=0.95\linewidth]{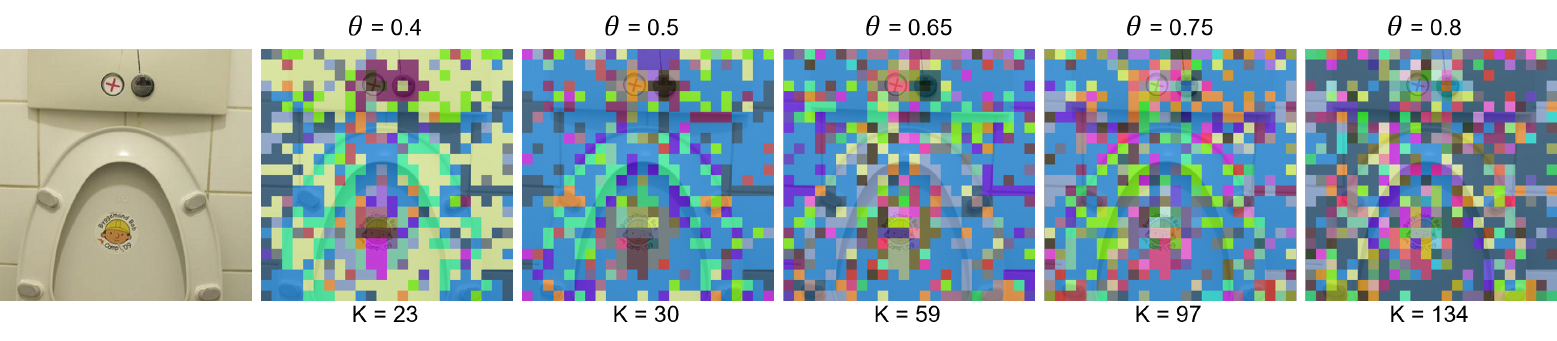}
    \includegraphics[width=0.95\linewidth]{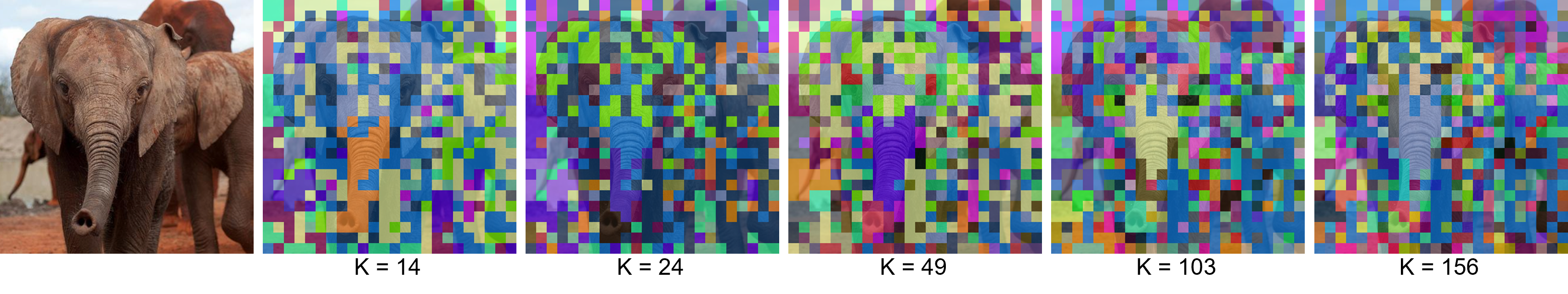}
    \includegraphics[width=0.95\linewidth]{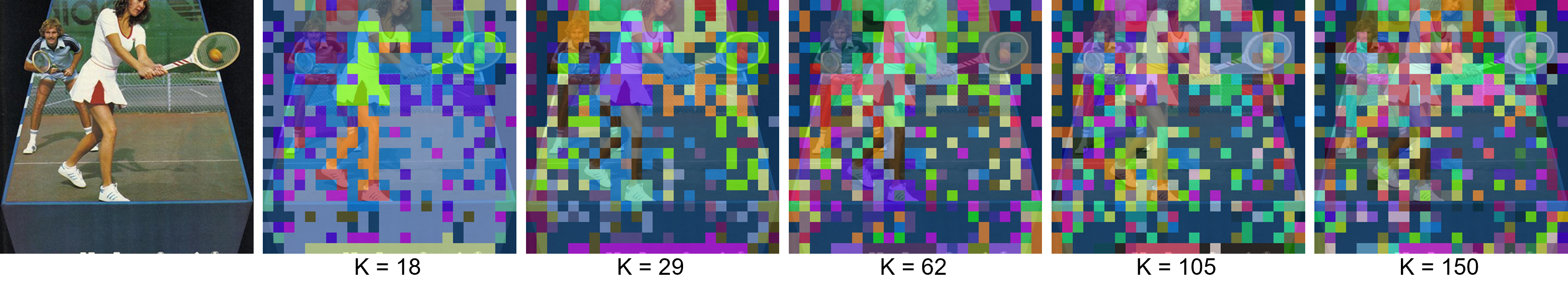}
    \includegraphics[width=0.95\linewidth]{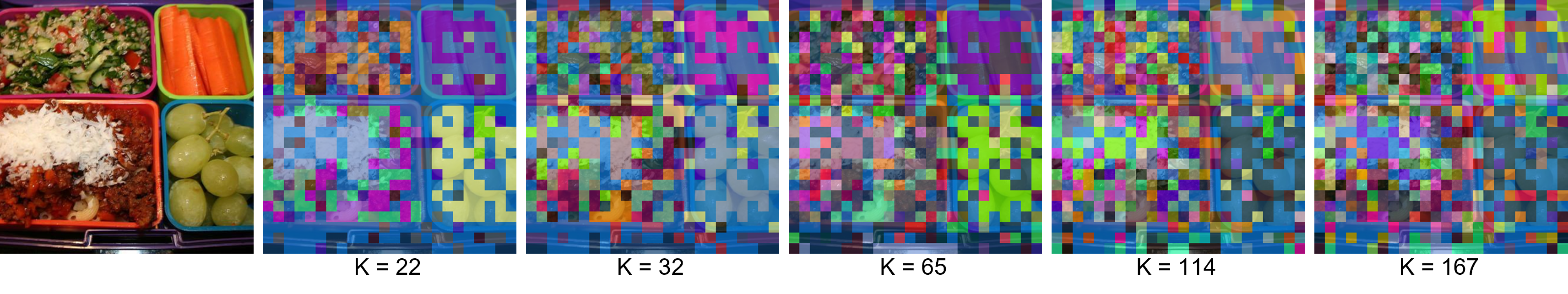}
    \includegraphics[width=0.95\linewidth]{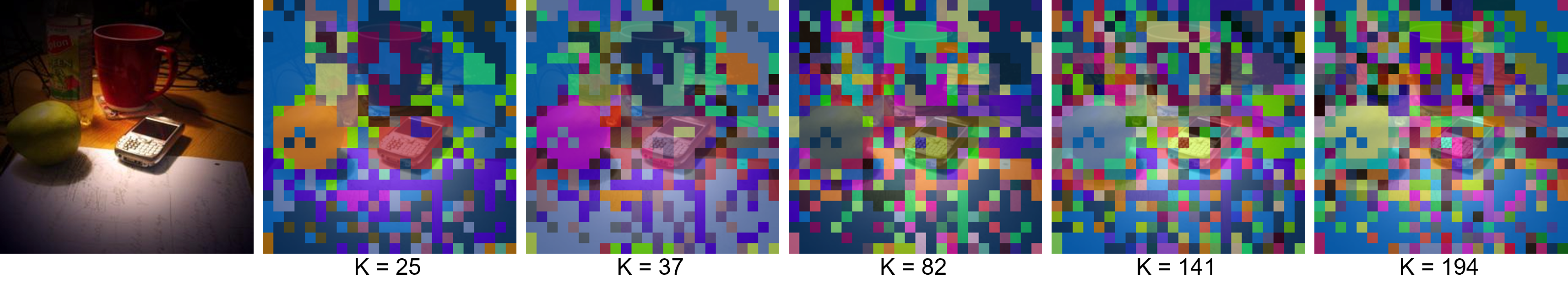}
    \includegraphics[width=0.95\linewidth]{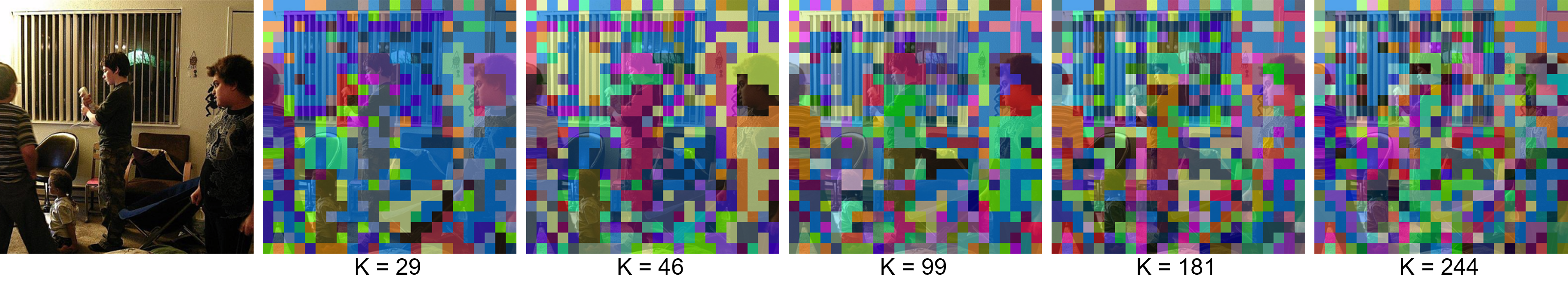}
    \includegraphics[width=0.95\linewidth]{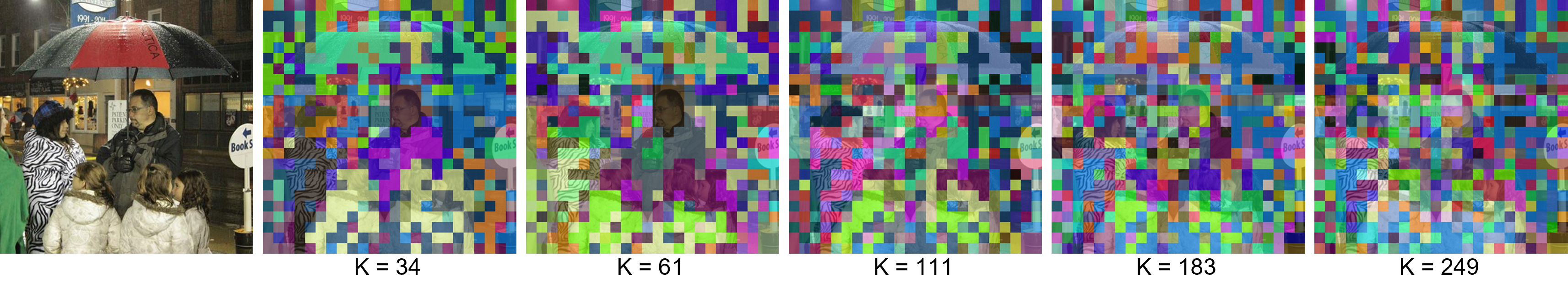}
    \caption{\textbf{Additional cluster visualizations produced by \method.}}
    \label{fig:extra_clustering}
\end{figure*}

We provide additional qualitative examples in \cref{fig:extra_clustering} that illustrates how \method\ forms semantically coherent clusters across diverse scenes.
Each example assigns a distinct color to patches belonging to the same cluster, allowing us to inspect how the feature-space grouping translates into spatial regions in the original image.
As in the main manuscript, the number of clusters is determined dynamically based on the image content, and the resulting visual patterns clearly reflect this adaptivity; that is, relatively simpler scenes yield compact clusters with large spatial support, while more complex or cluttered scenes produce a larger number of fine-grained clusters.
Varying the similarity threshold $\theta$ also produces the expected behavior, where a higher value leads to a more fragmented grouping.

These examples highlight that \method\ consistently discovers semantic units such as objects, parts, and salient regions without any pixel-level annotation, segmentation masks, or bounding box supervision.
Clusters formed purely from feature similarity often align with intuitive semantic boundaries, illustrating the effectiveness of our disentanglement mechanism.

\section{Analysis on Resulting Token Counts}
\label{appendix:appendix_token_stats}

\begin{table*}[t]
    \small
    \centering
    \resizebox{\textwidth}{!}{%
    \begin{tabular}{l * {11}{c}}
        \toprule
        \textbf{$\theta$} & \textbf{Pretrain} & \textbf{Finetune} & \textbf{MMB} & \textbf{VQA\textsuperscript{v2}} & \textbf{GQA} & \textbf{MME} & \textbf{MM-Vet} & \textbf{VQA\textsuperscript{Text}} & \textbf{SQA\textsuperscript{IMG}} & \textbf{POPE} & \textbf{Avg.} \\
        \midrule
        0.3  & 14.8 $\pm$ 8.4   & 14.9 $\pm$ 5.3  & 11.2 $\pm$ 4.3  & 13.3 $\pm$ 3.9  & 14.6 $\pm$ 3.7  & 12.6 $\pm$ 5.1  & 15.4 $\pm$ 7.3  & 18.3 $\pm$ 8.1  & 10.6 $\pm$ 5.0  & 14.3 $\pm$ 3.9  & 13.5 $\pm$ 4.3  \\
        0.4  & 22.1 $\pm$ 11.7  & 23.7 $\pm$ 7.9  & 18.0 $\pm$ 6.5  & 22.0 $\pm$ 6.4  & 24.7 $\pm$ 6.2  & 20.5 $\pm$ 7.2  & 23.0 $\pm$ 9.9  & 28.7 $\pm$ 11.8 & 16.2 $\pm$ 6.5  & 24.0 $\pm$ 6.2  & 22.4 $\pm$ 6.9  \\
        0.5  & 32.4 $\pm$ 15.7  & 37.2 $\pm$ 12.2 & 27.1 $\pm$ 10.9 & 35.1 $\pm$ 10.3 & 40.0 $\pm$ 10.5 & 32.5 $\pm$ 11.4 & 33.0 $\pm$ 13.2 & 45.8 $\pm$ 19.4 & 22.4 $\pm$ 8.1  & 39.3 $\pm$ 11.8 & 35.7 $\pm$ 11.4 \\
        0.62 & 54.9 $\pm$ 23.4  & 66.0 $\pm$ 18.8 & 50.0 $\pm$ 17.6 & 63.1 $\pm$ 18.3 & 70.1 $\pm$ 17.1 & 59.1 $\pm$ 18.3 & 57.1 $\pm$ 18.1 & 72.8 $\pm$ 24.7 & 41.6 $\pm$ 13.4 & 68.7 $\pm$ 18.2 & 63.7 $\pm$ 18.9 \\
        0.65 & 62.3 $\pm$ 26.0  & 76.5 $\pm$ 24.7 & 58.3 $\pm$ 20.5 & 73.5 $\pm$ 21.1 & 81.3 $\pm$ 19.3 & 69.3 $\pm$ 21.3 & 65.2 $\pm$ 19.8 & 83.4 $\pm$ 27.7 & 48.3 $\pm$ 15.7 & 80.1 $\pm$ 21.0 & 74.1 $\pm$ 21.8 \\
        0.75 & 110.3 $\pm$ 42.0 & 138.8 $\pm$ 41.2& 108.4$\pm$ 42.0 & 136.2$\pm$ 38.5 & 146.2$\pm$ 33.8 & 133.2$\pm$ 44.8 & 114.2$\pm$ 36.1 & 145.8$\pm$ 47.3 & 88.5 $\pm$ 31.1 & 147.1$\pm$ 39.2 & 136.5$\pm$ 39.6 \\
        \bottomrule
    \end{tabular}
    }
    \caption{
    \textbf{Resulting token counts of our proposed method across datasets for multiple thresholds}. Reported as mean $\pm$ standard deviation. The Avg. column averages over evaluation benchmarks only.}
    \label{tab:token_stats}
\end{table*}

\cref{tab:token_per_benchmark} of the main manuscript reports the average number of tokens produced at $\theta = 0.65$.
In this section, we extend this by providing mean and standard deviation statistics across multiple thresholds.
These measurements are computed over all images within each benchmark, illustrating how \method\ adapts its token budget depending on both the contents of input images and the similarity threshold.


In \cref{tab:token_stats}, we observe clear and consistent trends of the resulting token counts across thresholds and datasets.
A lower threshold such as $\theta = 0.4$ yields compact token sets with small variance, as visually dominant regions are merged into broader clusters.
Increasing $\theta$ makes the clustering more selective, producing more finer-grained tokens and higher token-count variance, particularly on benchmarks containing text, cluttered objects, or complex compositions (\textit{e.g.}, TextVQA or POPE).
Conversely, datasets with simpler scenes, such as SQA-IMG, maintain a narrow token-count range across all thresholds.

Collectively, \method\ adjusts its token budget according to the inherent visual complexity of each image rather than relying on a fixed grid-based reduction.
The resulting distribution of token counts demonstrates that DiVT responds naturally to semantic density, enabling compute-efficient representations without sacrificing expressiveness.

\end{document}